\relax
\documentclass[a4]{article}
\usepackage[a4paper,left=0.5in,right=0.5in,top=0.5in,bottom=0.5in,%
footskip=.25in]{geometry}
\usepackage{titling}
\usepackage{helvet}

\usepackage[hyphens]{url}
\usepackage{graphicx}
\usepackage{caption}
\usepackage[superscript]{cite}
\usepackage{enumitem}%
\usepackage[justification=centering]{caption}

\usepackage{stfloats}
\usepackage[hang,flushmargin, norule]{footmisc}
\usepackage{marvosym}

\usepackage{lineno}

\usepackage[dvipsnames]{xcolor}

\usepackage{titlesec}
\usepackage{multicol}
\usepackage{cite}
\usepackage{amsmath,amssymb,amsfonts}
\usepackage{algcompatible}
\usepackage{algorithm,algpseudocode}
\usepackage{graphicx}
\usepackage{booktabs, multirow, soul}
\usepackage[flushleft]{threeparttable}
\usepackage{xspace, mathtools, comment}
\usepackage{caption}
\usepackage{subcaption}
\usepackage[resetlabels,labeled]{multibib}
\usepackage{etoolbox}

\usepackage{multicol}
\usepackage{lipsum}

\usepackage{placeins}

\usepackage{xr}
\titlespacing*{\section}{0pt}{1\baselineskip}{0.1\baselineskip}
\titlespacing*{\subsection}{0pt}{1\baselineskip}{0.1\baselineskip}
\titlespacing*{\subsubsection}{0pt}{1\baselineskip}{0.1\baselineskip}

\titleformat*{\section}{\normalfont\sffamily\large\bfseries}
\titleformat*{\subsection}{\normalfont\sffamily\normalsize\bfseries}
\titleformat*{\subsubsection}{\normalfont\sffamily\small\bfseries}

\algnewcommand{\LineComment}[1]{\Statex \(\triangleright\) #1}
\algnewcommand{\ShortLineComment}[1]{\Statex \hspace{1.8em}\(\triangleright\) #1}
\algnewcommand{\ShortShortLineComment}[1]{\Statex \hspace{3.1em}\(\triangleright\) #1}
\algnewcommand{\ShortShortShortLineComment}[1]{\Statex \hspace{4.4em}\(\triangleright\) #1}

\DeclareCaptionLabelSeparator{bar}{\space\textbar\space}
\renewcommand{\figurename}{Figure}
\DeclareCaptionSubType{subfigure}
\makeatletter
\renewcommand\@biblabel[1]{#1.}
\makeatother

\makeatletter
\renewenvironment{table*}%
{\renewcommand\familydefault\sfdefault
	\@float{table}}
{\end@float}
\makeatother

\makeatletter
\renewcommand{\scriptsize}{\@setfontsize\scriptsize{7}{7}}
\renewcommand{\footnotesize}{\@setfontsize\footnotesize{8}{9}}
\makeatother

\patchcmd{\thebibliography}
{\settowidth}
{\setlength{\parsep}{0pt}\setlength{\itemsep}{0pt plus 0pt}\settowidth}
{}{}
\apptocmd{\thebibliography}
{\small}
{}{}

\renewcommand{\footnoterule}{%
	\kern -3pt
	\hrule width 0.5\textwidth height 0.5pt
	\kern 2pt
}

\newcommand{\etal}{\mbox{\emph{et al.}}\xspace}

\newcommand{\ours}{\mbox{$\mathsf{G^2Retro}$}\xspace}
\newcommand{\oursb}{\mbox{$\mathsf{G^2Retro}$\text{-}$\mathsf{B}$}\xspace}
\newcommand{\oursens}{\mbox{$\mathsf{G^2Retro}$\text{-}$\mathsf{ens}$}\xspace}

\newcommand{\brics}{\mbox{BRICS}\xspace}

\newcommand{\graph}{\mbox{$\mathcal{G}$}\xspace}
\newcommand{\graphmm}{\mbox{$\mathcal{G}^M$}\xspace}
\newcommand{\graphmp}{\mbox{$\mathcal{G}^M_p$}\xspace}
\newcommand{\graphmr}{\mbox{$\mathcal{G}^M_r$}\xspace}
\newcommand{\graphms}{\mbox{$\mathcal{G}^M_s$}\xspace}
\newcommand{\graphm}{\mbox{$\mathcal{G}^{*}$}\xspace}
\newcommand{\graphp}{\mbox{$\mathcal{G}_p$}\xspace}
\newcommand{\graphr}{\mbox{$\mathcal{G}_r$}\xspace}
\newcommand{\graphs}{\mbox{$\mathcal{G}_s$}\xspace}
\newcommand{\graphsi}{\mbox{$\mathcal{G}_{s,i}$}\xspace}

\newcommand{\atoms}{\mbox{$\mathcal{A}$}\xspace}
\newcommand{\bonds}{\mbox{$\mathcal{B}$}\xspace}
\newcommand{\vertices}{\mbox{$\mathcal{V}$}\xspace}

\newcommand{\edges}{\mbox{$\mathcal{E}$}\xspace}

\newcommand{\neigh}{\mbox{$\mathcal{N}$}\xspace}

\newcommand{\vocab}{\mbox{$\mathcal{Z}$}\xspace}

\newcommand{\neighCSet}{\mbox{$\mathcal{C}_{\mathsf{BF}}$}\xspace}
\newcommand{\neighACSet}{\mbox{$\mathcal{C}_{\mathsf{A}}$}\xspace}

\newcommand{\bondfmC}{\mbox{$\mathsf{BF}$-$\mathsf{center}$}\xspace}
\newcommand{\bondcgC}{\mbox{$\mathsf{BC}$-$\mathsf{center}$}\xspace}
\newcommand{\atomC}{\mbox{$\mathsf{A}$-$\mathsf{center}$}\xspace}

\newcommand{\atom}{\mbox{$a$}\xspace}

\newcommand{\atomEmb}{\mbox{$\mathbf{a}$}\xspace}

\newcommand{\bond}{\mbox{$b$}\xspace}

\newcommand{\bondEmb}{\mbox{$\mathbf{b}$}\xspace}

\newcommand{\rcenter}{\mbox{$C$}\xspace}
\newcommand{\score}{\mbox{$s$}\xspace}

\newcommand{\node}{\mbox{$n$}\xspace}

\newcommand{\nodeEmb}{\mbox{$\mathbf{n}$}\xspace}

\newcommand{\edge}{\mbox{$e$}\xspace}
\newcommand{\edgeEmb}{\mbox{$\mathbf{e}$}\xspace}

\newcommand{\frag}{\mbox{$z$}\xspace}

\newcommand{\centerEmb}{\mbox{$\mathbf{c}$}\xspace}
\newcommand{\hidden}{\mbox{$\vect{h}$}\xspace}

\newcommand{\MPN}{\mbox{$\mathsf{MPN}$}\xspace}
\newcommand{\GMPN}{\mbox{$\mathsf{GMPN}$}\xspace}
\newcommand{\BMPN}{\mbox{$\mathsf{FMPN}$}\xspace}
\newcommand{\mess}{\mbox{$\mathbf{m}$}\xspace}
\newcommand{\mol}{\mbox{$M$}\xspace}
\newcommand{\molp}{\mbox{$M_p$}\xspace}
\newcommand{\molr}{\mbox{$M_r$}\xspace}
\newcommand{\mols}{\mbox{$M_s$}\xspace}
\newcommand{\molm}{\mbox{$M^*$}\xspace}

\newcommand{\RCI}{\mbox{$\mathsf{RCI}$}\xspace}

\newcommand{\syncomp}{\mbox{$\mathsf{SC}$}\xspace}

\newcommand{\ACP}{\mbox{$\mathsf{ACP}$}\xspace}
\newcommand{\AAP}{\mbox{$\mathsf{AAP}$}\xspace}
\newcommand{\AACP}{\mbox{$\mathsf{AACP}$}\xspace}
\newcommand{\AATP}{\mbox{$\mathsf{AATP}$}\xspace}
\newcommand{\BTCP}{\mbox{$\mathsf{BTCP}$}\xspace}

\newcommand{\pstrans}{\mbox{$p$2$s$-$\mathtt{T}$}\xspace}

\newcommand{\jtvae}{\mbox{$\mathsf{JT}$-$\mathsf{VAE}$}\xspace}
\newcommand{\modof}{\mbox{$\mathsf{Modof}$}\xspace}

\newcommand{\retrosim}{\mbox{$\mathsf{Retrosim}$}\xspace}
\newcommand{\neuralsim}{\mbox{$\mathsf{Neuralsym}$}\xspace}
\newcommand{\gln}{\mbox{$\mathsf{GLN}$}\xspace}
\newcommand{\mhnreact}{\mbox{$\mathsf{MHNreact}$}\xspace}
\newcommand{\localretro}{\mbox{$\mathsf{LocalRetro}$}\xspace}
\newcommand{\scrop}{\mbox{$\mathsf{SCROP}$}\xspace}
\newcommand{\lvtrans}{\mbox{$\mathsf{LV}$-$\mathsf{Trans}$}\xspace}
\newcommand{\augtrans}{\mbox{$\mathsf{AT}$}\xspace}
\newcommand{\gtos}{\mbox{$\mathsf{Graph2SMILES}$}\xspace}
\newcommand{\dual}{\mbox{$\mathsf{Dual}$}\xspace}
\newcommand{\get}{\mbox{$\mathsf{GET}$}\xspace}
\newcommand{\chemformer}{\mbox{$\mathsf{Chemformer}$}\xspace}
\newcommand{\retroformer}{\mbox{$\mathsf{Retroformer}$}\xspace}
\newcommand{\tiedformer}{\mbox{$\mathsf{Tied Transformer}$}\xspace}
\newcommand{\gta}{\mbox{$\mathsf{GTA}$}\xspace}
\newcommand{\rsmiles}{\mbox{$\mathsf{R{\textendash}SMILES}$}\xspace}
\newcommand{\retroxpert}{\mbox{$\mathsf{RetroXpert}$}\xspace}
\newcommand{\gtog}{\mbox{$\mathsf{G2G}$}\xspace}
\newcommand{\retroprime}{\mbox{$\mathsf{RetroPrime}$}\xspace}
\newcommand{\graphretro}{\mbox{$\mathsf{GraphRetro}$}\xspace}

\newcommand{\megan}{\mbox{$\mathsf{MEGAN}$}\xspace}
\newcommand{\tree}{\mbox{$\mathcal{G}^B$}\xspace}
\newcommand{\treep}{\mbox{$\mathcal{G}^B_p$}\xspace}

\newcommand{\HRD}{\mbox{$\mathsf{HRD}$}\xspace}
\newcommand{\LRD}{\mbox{$\mathsf{LRD}$}\xspace}

\newcommand{\tb}{\mbox{TB}\xspace}
\newcommand{\tf}{\mbox{TF}\xspace}
\newcommand{\semitb}{\mbox{Semi-TB}\xspace}

\newcommand{\vect}[1]{\mathbf{#1}}

\newcommand{\relu}{\mbox{$\text{ReLU}$}\xspace}

\title{\ours as a Two-Step Graph Generative Models for Retrosynthesis Prediction}
\date{\vspace{-5ex}}

\author{
	Ziqi Chen\textsuperscript{\rm 1}, 
	Oluwatosin R. Ayinde\textsuperscript{\rm 2},
	James R. Fuchs\textsuperscript{\rm 2},
	Huan Sun\textsuperscript{\rm 1,3},
	Xia Ning\textsuperscript{\rm 1,3,4 \Letter}\\
}
\begin{document}

\maketitle

\begin{center}
\normalfont\sffamily\large\bfseries{Abstract}
\end{center}

Retrosynthesis is a procedure where a target molecule is transformed into potential reactants 
and thus the synthesis routes can be identified. 
{Recently, computational approaches have been developed to accelerate the design of synthesis routes.}
{In this paper, }we develop a generative framework \ours for one-step retrosynthesis prediction.
\ours imitates the reversed logic of synthetic reactions. 
It first predicts the reaction centers in the target molecules (products), identifies the synthons needed to assemble 
the products, and transforms these synthons into reactants.  
\ours defines a comprehensive set of reaction center types, and  
learns from the molecular graphs of the products to 
predict potential reaction centers.  
To complete synthons into reactants, 
\ours considers all the involved synthon structures and the product structures 
to identify the optimal completion paths, 
and accordingly attaches small substructures sequentially 
to the synthons. 
Here we show that \ours is able to better predict the reactants for given products 
in the benchmark dataset than the state-of-the-art methods.

\section*{Introduction}

Retrosynthesis is a procedure where a target molecule is transformed into potential reactants 
and thus the synthesis routes can be identified. 
One-step retrosynthesis, which transforms a molecule into the possible direct reactants that can be used 
to synthesize the molecule, serves as the foundation of multi-step synthesis planning~{\cite{Segler2018,chen2020}}
that identifies a full synthesis route in which the target molecule can be made through a series of one-step synthesis reactions.  
In drug discovery, identifying feasible synthesis routes 
for drug-like molecules remains a factor that substantially challenges medicinal chemists in making the 
desired molecules experimentally{~\cite{Blakemore2018}}.
An extensive, diverse library of high-quality synthesis routes for a given molecule has the potential to
enable more feasible reaction solutions
starting from commercially available, chemical building blocks, 
and to provide more options for operationally simple, high-yielding 
transformations using widely accessible reactants. 

Current retrosynthesis planning is primarily conducted by synthetic and medicinal chemists based on 
their knowledge and experience. 
It has been long known that there exists substantial disagreement among chemists in assessing 
synthesisbilty and designing synthesis routes~{\cite{Lajiness2004,Huang2011,Takaoka2003, Kutchukian2012}}. 
In addition, an ever-increasing number of new chemical reactions makes it highly challenging for a chemist to keep up to date. 
Therefore, a data-driven model that predicts synthetic reactions could provide a useful complement to chemist evaluations, 
and could provide a large pool of potential reactions that the chemists can consider. 
There exist proprietary synthesis reaction databases manually curated from the literature, including Reaxys~\cite{reaxys} and
SciFinder~\cite{scifinder}.
Unfortunately, the high prices of these databases act to limit their accessibility in some academic and small biotech settings. 
Open-sourced synthesis reaction databases such as the Open Reaction Database~\cite{ord}
are limited in the reactions they cover (e.g., majorities are United States Patent and Trademark Office (USPTO) public reactions~\cite{uspto}) and their search functionalities (e.g., 
via SMILES strings).
Even with the aid of these databases, the development of new reactions and synthetic pathways for the preparation of 
challenging molecules remains non-trivial. 
In addition, database searches can be time-consuming with low throughput, particularly when without extensive domain knowledge 
to guide the process.
Recent \emph{in silico} retrosynthesis prediction methods using deep learning
~\cite{Coley2017,Segler2017,dai2019,seidl2021modern,chen2019localretro,
		Zheng2019,Chen2019,Mao2021,Irwin2022,Tu2021,Kim2021,seo2021gta,sun2021towards,Wan2022,Sacha_2021,
		yan2020retroxpert,shi2020g2g,somnath2021learning,Wang2021,Tetko_2020,Zhong2022} 
have enabled alternative computationally generative processes to accelerate the conventional paradigm. 
These deep-learning methods learn from string-based representations
(SMILES) or graph representations of given molecules, and generate possible reactant structures that can be used to synthesize these 
molecules, leveraging the advancement of natural language processing~\cite{Vaswani2017}, 
graph neural networks~{\cite{kipf2017}}, variational auto-encoders~{\cite{kinma2013}} and other techniques in deep learning. 
They have demonstrated strong potential to substantially accelerate and advance retrosynthesis analysis{~\cite{Venkatasubramanian2022}}. 
In this manuscript, 
we focus on the one-step retrosynthesis prediction, which predicts the possible direct reactants for the synthesis 
of the target molecules,
and acts as the foundation of multi-step retrosynthesis analysis~{\cite{Segler2018}}.

We develop a semi-template-based method via deep learning for one-step retrosynthesis prediction, denoted as \ours. 
\ours imitates the reversed logic of synthetic reactions: 
it first predicts the reaction centers in the target molecules, identifies the synthons needed to assemble the final products, and transforms these synthons into reactants. 
Therefore, \ours
follows the semi-template-based frame, as in the previous methods~\cite{shi2020g2g,yan2020retroxpert,somnath2021learning,Wang2021}.
To predict reaction centers, \ours learns from the molecular graphs of the products via a customized graph representation learning~\cite{Chen2020survey} and embedding approach ({in {``Molecule Representation Learning''} Section}), 
and uses the graph structures to predict potential reaction centers. 
\ours defines a comprehensive set of reaction center types, and for each reaction center type, 
uses the graph structures that are most relevant to that reaction center type ({in ``Reaction Center Identification'' Section}). 
\oursb integrates information of synthetically accessible fragments in its molecule graph representation learning 
({in Supplementary Note 1).

The predicted reaction centers by \ours split the products into synthons. 
To complete synthons into reactants, \ours considers all the involved synthon structures and the product structures 
to identify the optimal completion paths ({in ``Attachment Continuity Prediction (\AACP)'' Section}), 
and accordingly attaches small substructures (i.e., bonds or rings) sequentially 
to the synthons until the extended synthon structures are predicted as possible reactants ({in ``Attachment Type Prediction (\AATP)'' Section}).
All the involved predictions in \ours and \oursb are done via tailored neural networks. 
Note that \ours and \oursb allow multiple reaction centers and multiple completion paths for each product to increase diversity in its 
predicted reactions. 
That is, the top predicted reaction centers (according to predicted likelihoods) are all tested in synthon completion to produce 
different reactions. 
Meanwhile, to avoid the exhaustive generation of all possible reactions from the top reaction centers, 
\ours prioritizes the most possible completion paths via a new beam search strategy ({in ``Inference'' Section}). 
An ensemble of \ours was also developed, denoted as \oursens, an ensemble of \ours, increases the pool of generated reactions by combining multiple 
\ours models and their predictions. 
Figure~\ref{fig:model} presents an overview of \ours. 
{A comprehensive review of existing retrosynthesis prediction methods and related fragment-based molecular generation methods 
is available in ``Related Work'' Section.}

As a summary, \ours has the following advantages:
\begin{itemize}[leftmargin=*]
\item {{\ours} follows a semi-template-based framework,
predicts reaction centers of different types in products first, and then transforms the resulting synthons 
into reactants by adding substructures to the synthons. 
This process imitates the reversed logic of synthetic reactions and enables necessary interpretability as to 
which reaction centers are predicted by \ours, which reactants are generated from the reaction centers and 
the corresponding step-by-step generation process.}
\item {\ours defines a comprehensive set of reaction center types, covering 97.5\% of the test data 
and conforming to synthetic chemistry knowledge. 
New customized neural networks are developed to predict each type of the reaction centers as well as 
their associated atom changes. 
Multiple reaction center candidates are considered for each product to enable diverse reactions generated from different reaction 
centers in the predicted reactions. }
\item {{\ours} develops a new fragment-based generation strategy {compared to the previous semi-template-based methods~\cite{shi2020g2g,yan2020retroxpert,somnath2021learning,Wang2021},} to complete synthons into reactants by 
sequentially attaching substructures (i.e., bonds and rings) starting from the predicted reaction centers {(in ``Synthon Completion'' Section)}.  
The prediction of these substructure attachments utilizes a holistic view of 
the most updated structures of the synthon to be completed, and the structures of the final product and other synthons.}
\item {\ours employs a new, effective beam search strategy {compared to the previous semi-template-based methods~\cite{shi2020g2g,yan2020retroxpert,somnath2021learning,Wang2021},} that prioritizes the most possible reactants and the corresponding 
completion actions along the synthon completion paths. The beam search also allows multiple different reaction centers, enabling 
diversity in the completed reactants. }
\item {\ours and \oursb are compared with nineteen baseline methods 
and demonstrate the state-of-the-art performance over the benchmark data {(in ``Overall Comparison'' Section)}. 
Case studies show that \ours could propose {diverse and reasonable} synthesis routes with high predicted likelihoods that are not included in the benchmark data {(in ``Case Study'' Section)}.}
\item {{\oursens} is an ensemble of {\ours} models 
and demonstrates strong performance on the benchmark data compared to two baseline methods with data augmentation {(in ``Performance of Ensemble-based Methods'' Section)}.} 
\end{itemize}

\begin{figure*}[h]
	\centering
	\includegraphics[width=0.9\linewidth]{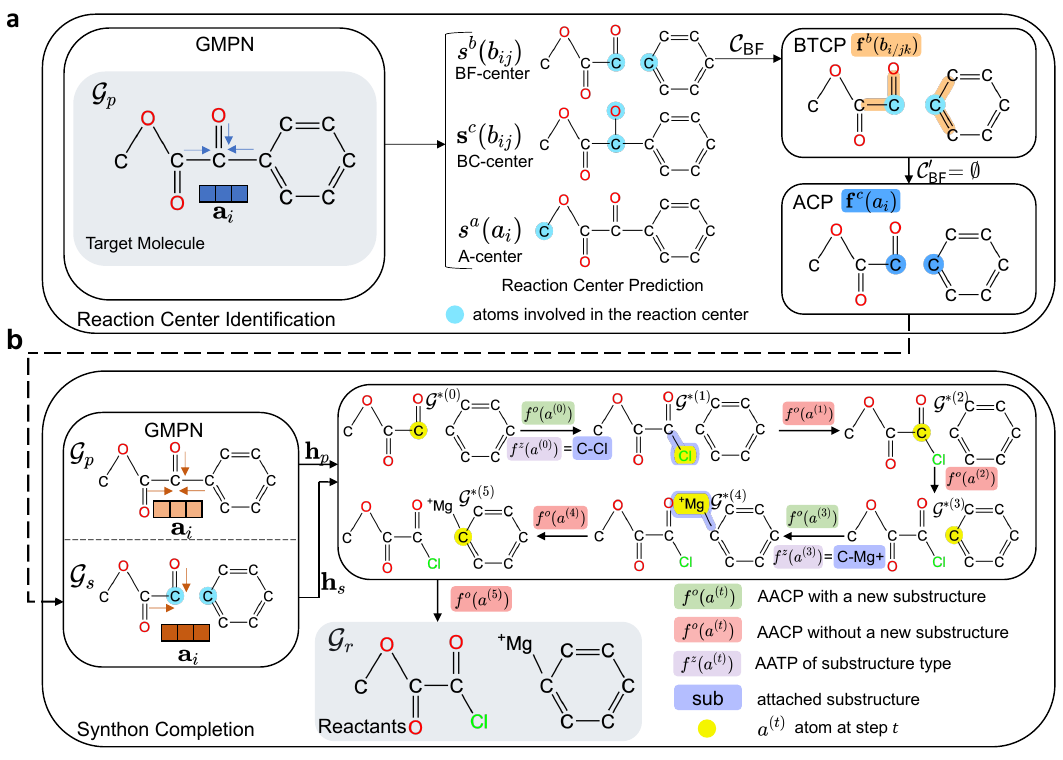}
	\caption{\textbf{\ours retrosynthesis prediction process.} 
	\textbf{a} \ours reaction center identification. \ours uses 
	a graph message passing network (\GMPN); 
	\ours predicts three types of reaction centers: newly formed bonds (\bondfmC), 
	bonds with type changes (\bondcgC), and atoms with leaving fragments (\atomC); 
	for \bondfmC, \ours also predicts bonds that have type changes induced by the newly formed bonds 
	(\BTCP); for all the reaction center types, \ours predicts atoms with charge changes (\ACP). 
	 \textbf{b} \ours synthon completion. \ours uses \GMPN to represent both the products and the synthons; 
	 \ours sequentially predicts whether a new substructure should be attached (\AACP) and the type of the attachment (\AATP); 
	 \ours adds predicted substructures until \AACP predicts `stop'.}
	\label{fig:model}
\end{figure*}

\section*{Results}
\label{sec:results}

\subsection*{Overall Comparison}
\label{sec:results:overall}
%


\begin{table*}[h]
	\captionof{table}{\textbf{Overall comparison {on retrosynthesis prediction} in top-$k$ accuracy (\%)}}
	\label{tbl:overall}
  \vspace{-8pt}    
\begin{center}
\begin{small}
\begin{threeparttable}
	\begin{tabular}{
			@{\hspace{2pt}}l@{\hspace{5pt}}
			@{\hspace{2pt}}l@{\hspace{5pt}}
			@{\hspace{2pt}}r@{\hspace{5pt}}
			@{\hspace{5pt}}c@{\hspace{5pt}}
			@{\hspace{5pt}}c@{\hspace{5pt}}
			@{\hspace{5pt}}c@{\hspace{5pt}}
			@{\hspace{5pt}}c@{\hspace{5pt}}
			@{\hspace{2pt}}r@{\hspace{2pt}}
			@{\hspace{5pt}}c@{\hspace{5pt}}
			@{\hspace{5pt}}c@{\hspace{5pt}}
			@{\hspace{5pt}}c@{\hspace{5pt}}
			@{\hspace{5pt}}c@{\hspace{5pt}}
		}
			\toprule
			\multirow{2}{*}{\parbox{0.05\textwidth}{Method type}} & \multirow{2}{*}{Method} & \multirow{2}{*}{Coverage(\%)} 
			& \multicolumn{4}{c}{Reaction type known} & & \multicolumn{4}{c}{Reaction type unknown}\\
			\cmidrule(lr){4-7} \cmidrule(lr){9-12}
			& & & 1 & 3 & 5 & 10 & &  1 & 3 & 5 & 10 \\	
			\midrule
			\multirow{5}{*}{\parbox{0.08\textwidth}{{\tb}}}
			 & \retrosim{{\cite{Coley2017}}}   & 100.0 & 52.9 & 73.8 & 81.2 & 88.1 & & 37.3 & 54.7 & 63.3 & 74.1 \\
			 & \neuralsim{{\cite{Segler2017}}}  & 100.0 & 55.3 & 76.0 & 81.4 & 85.1 & & 44.4 & 65.3 & 72.4 & 78.9 \\
			 & \gln{{\cite{dai2019}}}      & 93.3 & \textbf{64.2} & 79.1 & 85.2 & 90.0 & & 52.5 & 69.0 & 75.6 & 83.7 \\
			 & \mhnreact{{\cite{seidl2021modern}}} & 100.0 & -   &  -   &   -  &   -  & & 50.5 & 73.9 & 81.0 & 87.9 \\
			 & \localretro{{\cite{chen2019localretro}}} &  98.1 & 63.9 & \textbf{86.8} & \textbf{92.4} & \textbf{96.3} & & \textbf{53.4} & \textbf{77.5} & \textbf{85.9} & \textbf{92.4} \\
			 \midrule
			 \multirow{10}{*}{\parbox{0.08\textwidth}{{\tf}}}
			 & \scrop{{\cite{Zheng2019}}}          & \multirow{10}{*}{100.0} & 59.0 & 74.8 & 78.1 & 81.1 & & 43.7 & 60.0 & 65.2 & 68.7 \\
			 & \lvtrans{{\cite{Chen2019}}}          &  & -   &  -   &  -   &  -   & & 40.5 & 65.1 & 72.8 & 79.4 \\
			 & \get{{\cite{Mao2021}}}   	            &  & 57.4 & 71.3 & 74.8 & 77.4 & & 44.9 & 58.8 & 62.4 & 65.9 \\
			 & \chemformer{{\cite{Irwin2022}}}   &  &  -  &   -  &   -  &  -   & & \textbf{54.3} &   -  & 62.3 & 63.0 \\
			 & \gtos{{\cite{Tu2021}}}                   &  &  - &   -  &   -  &  -   & & 51.2 & 66.3 & 70.4 & 73.9 \\
			 & \tiedformer{{\cite{Kim2021}}}        &  &  - &   -  &   -  &  -   & & 47.1 & 67.1 & 73.1 & 76.3 \\
			 & \gta{{\cite{seo2021gta}}}               &  &  - &   -  &   -  &  -   & & 51.1 & 67.6 & 74.8 & 81.6 \\
			 & \dual{{\cite{sun2021towards}}}      &  &  \textbf{65.7} & 81.9 & 84.7 & 85.9 & & 53.6 & 70.7 & 74.6 & 77.0 \\
			 & \retroformer{{\cite{Wan2022}}}      &  & 64.0  & \textbf{82.5} & 86.7 & 90.2 & & 53.2 & \textbf{71.1} & 76.6 & 82.1 \\
			 & \megan{{\cite{Sacha_2021}}}        &  & 60.7 & 82.0 & \textbf{87.5} & \textbf{91.6} & & 48.1 & 70.7 & \textbf{78.4} & \textbf{86.1} \\
			 \midrule
			 \multirow{7}{*}{\parbox{0.08\textwidth}{{\semitb}}}
			 & \retroxpert{{\cite{yan2020retroxpert}}}   & 100.0 & 62.1 & 75.8 & 78.5 & 80.9 & & 50.4 & 61.1 & 62.3 & 63.4 \\
			 & \gtog{{\cite{shi2020g2g}}}      & 97.9 & 61.0 & 81.3 & 86.0 & 88.7 & & 48.9 & 67.6 & 72.5 & 75.5 \\
			 & \graphretro{{\cite{somnath2021learning}}}  & 95.0 &63.9 & 81.5 & 85.2 & 88.1 & & 53.7 & 68.3 & 72.2 & 75.5 \\
			 & {\retroprime}{{\cite{Wang2021}}} & 100.0 & \textbf{64.8} & 81.6 & 85.0 & 86.9 & & 51.4 & 70.8 & 74.0 & 76.1 \\
			 \cmidrule(lr){2-11}
			 & \ours & 97.5 & 63.1  &  \textbf{84.2}  & \textbf{88.5} &  \textbf{91.7}   & & \underline{53.9} & \textbf{74.6} & \underline{80.7} & \underline{86.6} \\
			 &  \oursb & 97.5 & 63.6  &  \underline{83.6}  &  \underline{88.4}  &  91.5   & & \textbf{54.1} & \underline{74.1} & \textbf{81.2} & \textbf{86.7} \\
			\bottomrule
	\end{tabular}%
		\begin{tablenotes}[normal,flushleft]
		\begin{footnotesize}
	\item \!Columns with 1, 3, 5 and 10 present top-1, top-3, top-5 and top-10 accuracies, respectively. 
	{Column ``Coverage(\%)'' represents the percentage of test reactions that the methods can be applied to.}
	Best top-$k$ accuracy values among the methods of each type are in \textbf{bold}.  
	Top-$k$ accuracy values of \ours and \oursb are \underline{underlined} if they are not the best but still better than all the baselines of the respective type.
	{All the baseline results are reported in their original papers, where ``-" represents that the corresponding results are 
	not reported. }
		\par 
		\end{footnotesize}
	\end{tablenotes}
\end{threeparttable}
\end{small}
  \vspace{-10pt}    
\end{center}
\end{table*}

Table~\ref{tbl:overall} presents the overall comparison between \ours, {{\oursb}} and the baseline methods on one-step retrosynthesis 
under two conditions, following the standard protocol in literature~\cite{Coley2017,Segler2017,dai2019,Mao2021,sun2021towards,Wang2021,yan2020retroxpert,shi2020g2g,somnath2021learning,Sacha_2021}: 
(1) when the reaction type is given \emph{a priori} for both model training
and inference (i.e., ``Reaction type known"); and (2) when the reaction type is always unknown (i.e., ``Reaction type unknown").
When the reaction type is known, \ours uses a one-hot encoder as an additional feature for each atom in product molecules
indicating the reaction type.  
Particularly, for \semitb methods, the 
performance in Table~\ref{tbl:overall} 
corresponds to the predictions out of the two steps, that is, the synthon completion is done according to a 
 reaction center that is \emph{predicted} from the reaction center prediction step.}
{Following the prior work~{\cite{Coley2017,dai2019,shi2020g2g,somnath2021learning}}, 
we used the top-$k$ ($k$=1,3,5,10) accuracy to evaluate the overall performance of all the methods.
Top-$k$ accuracy is the ratio of test products that have 
their ground truth correctly predicted among their top-$k$ predictions. Higher top-$k$ accuracy indicates better performance.}
{Note that ground truth reactions are those included in the benchmark data.
While there is always one ground-truth reaction for each product in the benchmark data, 
there may exist actually numerous feasible reactions for each product that are not included in the benchmark data. 
Therefore, reactants that are considered incorrect based on the benchmark data might still be plausible 
and included in other larger databases.}
%
%
%
{Also, n}ote that the top-$k$ accuracies of all the baseline methods are 
{the reported results in their original papers (issues related to the comparison among methods 
are discussed later).
{Details of baseline methods are available in ``Baselines'' Section.}
%

\subsubsection*{Comparison with semi-template-based (\semitb) methods}
\label{sec:results:overall:semi}

When the reaction type is known, compared to other \semitb methods, 
{\ours} achieves the best performance on top-3 (84.2\%), top-5 (88.5\%) and top-10 (91.7\%) accuracies,
{corresponding to 3.2\%, 2.9\%, and 3.4\% improvement over those from the best baselines 
(81.6\% for {\retroprime}~{\cite{Wang2021}} on top-3, 86.0\% and 88.7\% 
for {\gtog}~{\cite{shi2020g2g}} on top-5 and top-10)}
on these {three} metrics.
In terms of top-1 accuracy, {\oursb} achieves the third-best performance (63.6\%) compared to those of 
{\retroprime} (64.8\%) and \graphretro (63.9\%) on this metric.
While \ours underperforms \retroprime on one metric, it is {substantially} 
better than \retroprime on all the other metrics:
\ours outperforms \retroprime on top-3 accuracy at 3.2\%, on top-5 accuracy at 4.1\%, 
and on top-10 accuracy at 5.6\%.

When the reaction type is unknown, a similar trend is observed: \oursb outperforms all the \semitb
baseline methods on all the top accuracy metrics, with 0.7\% improvement over the best baseline \graphretro on top-1 accuracy, 
{
and 4.7\%, 9.7\% and 13.9\% improvement over those from the best baseline {\retroprime} on top-3, top-5 and top-10 accuracies.
}
\ours has a performance similar to that of {\oursb}, with an even better top-3 performance 74.6\% that is 5.4\%
improvement from that of \retroprime. 

Compared with the performance with known reaction types, all the methods including \ours and \oursb
have worse performance when the reaction types are unknown.
It is well-known in synthetic chemistry that there are several well-characterized reaction types. 
These types have distinct patterns in their reactions and reaction centers.
For example, acylation reactions are very common approaches to creating amide and sulfonamide linkages. 
They are known for their efficiency and high yields, especially when they involve acyl/sulfonyl halides~\cite{Brown_2015}. 
The improved performance with known reaction types integrated into retrosynthesis model training 
demonstrates that leveraging \emph{a priori} reaction type information could benefit retrosynthesis prediction in general. 
However, in real applications, reaction types are typically not available in retrosynthesis when only the target molecule is presented. 
The superior performance of \ours and \oursb in ``reaction type unknown" condition demonstrates their great utility in 
real applications. 

{As Table~\ref{tbl:overall} shows, {\ours} and {\oursb} can cover (i.e., can be applied to) 97.5\% of the test reactions, 
which determines the upper bound 
of accuracy values, due to the definition of reaction centers 
(the rest 2.5\% correspond to reactions with multiple newly formed or changed bonds).
Among other {\semitb} methods, {\gtog} and {\graphretro}{~{\cite{somnath2021learning}}} also have limited coverage on test set 
(97.9\% for {\gtog} and 95.0\% for {\graphretro}). \retroxpert{~{\cite{yan2020retroxpert}}} has 100\% coverage because its reactant 
SMILES generation from synthons recovers all possible reaction centers. 
\retroprime{~{\cite{Wang2021}}} also has 100\% coverage due to its very comprehensive set of reaction centers. 
Although \ours and \oursb cannot cover all possible cases in the test set, they still outperform other 
\semitb methods, measured over the entire test set. 
More discussion on the coverage of the two steps in {\semitb} methods is available
in the Section ``Individual Module Performance.''}


%
{{\graphretro} and {\retroprime} are two strong baselines.}
%
{\graphretro} has good top-1 accuracies but much worse results on other top accuracy metrics. 
According to its authors~\cite{somnath2021learning}, \graphretro tends to bias its beam search to the 
most possible reaction center. 
Thus, it may prioritize the most possible reactants from the most possible 
reaction center at the very top of its predictions. 
However, if the most possible reaction centers are not the ground truth, 
\graphretro would totally miss the ground truth in its beam search, resulting in poor performance on other top accuracy metrics. 
{In addition, such focused beam search limits the diversity of identified synthons, and thus the completed reactants.}
%
%
\retroprime achieves the best top-1 accuracy with reaction type known. 
It uses augmented SMILES strings (i.e., each product 
has multiple, equivalent, non-canonical SMILES strings) in training the two sequence-to-sequence transformers. 
It is likely that 
top results in \retroprime correspond to the ground truth but in different, augmented SMILES strings, and thus high top-1 accuracy but 
low and similar other top accuracies. 
These three \semitb baseline methods only perform well on one certain metric (in one certain condition), but do not show consistent optimality 
across many metrics or across the two conditions. 

Compared to these baselines, \ours always achieves the best performance on all the top accuracy metrics (except on top-1 accuracy when 
reaction types are known). 
High top-$k$ accuracies at all different $k$ are desired as they indicate the holistically high ranking positions of the ground truth in the 
predicted reactions, and thus the capability of models in recovering knowledge from data. 
High top-$k$ accuracies with $k>1$ may signify 
plausible reactions {not included in the dataset}, as will be examined later
in Section ``Case Study''. 
This is because high top-$k$ ($k>1$) accuracy implies that there might be a few 
reactions different from the ground truth but are very possible and thus are ranked on top.
{Such results may enable the exploration of multiple synthesis routes and may be of synthetic value if specific coupling methods fail or 
if specific starting materials are unavailable.}
%
%
From the above two aspects, over all the metrics, 
\ours and \oursb achieve the overall best performance compared to the three strong \semitb methods.
%


\oursb performs slightly better than \ours when the reaction types are unknown, but worse than \ours 
when the reaction types are known. \oursb integrates synthetically accessible fragments in atom 
embeddings (Equation S3 in Supplementary Note 1). 
When the reaction types are unknown, the fragment information provides additional local contexts to atoms, 
which could facilitate better decisions on reaction center prediction and synthon completion.
When the reaction types are known, atom embeddings directly integrate the reaction type information in \ours, 
which may outweigh the contextual information provided by the fragments, and thus \oursb does not achieve additional 
performance improvement from \ours.

\subsubsection*{Comparison with template-free (\tf) methods}
\label{sec:results:overall:free}

\ours and \oursb also demonstrate superior or competitive performance compared to \tf
methods on all the top accuracies. 
With reaction types known, \ours is the best on top-3, top-5 top-10 accuracies compared to 
all the template-free methods; with reaction types unknown, \oursb is the best on top-3, top-5 and 
top-10 accuracies, and is the second best one on top-1 accuracy.
For example, \ours is 4.9\% better than the best \tf method on top-3 accuracy (i.e., \retroformer~{\cite{Wan2022}}) with the reaction types unknown. 
Most \tf methods such as \dual~{\cite{sun2021towards}} and \chemformer~{\cite{Irwin2022}} have the competitive performance on top-1 accuracy
but relatively worse results on other top accuracy metrics. 
This could be due to that these \tf methods with SMILES representations may fail to generate diverse or even many valid reactants 
with beam search~\cite{Vijayakumar2016},
leading to limited variation in their predicted results, and thus low and similar 
top-$3$, top-$5$ and top-$10$ accuracies. 
This lack of diversity and richness in the predictions, in addition to the lack of interpretability during 
the chemical sequence transformation process, 
could hinder the application of \tf methods in retrosynthesis prediction. 
{However, the prediction diversity and richness in \ours is enabled by the multiple possible reaction centers predicted by \ours and 
the corresponding completed reactants. }
%

{In terms of the coverage on the test set, all the SMILES-based {\tf} methods  can cover the entire test set, 
because all the reactions can be represented as SMILES string transformation.
The graph-based {\tf} method {\megan}~{\cite{Sacha_2021}} also covers the entire test set due to its comprehensive set of graph edit
actions. 
{Compared to these {\tf} methods, though without the full coverage on the test set, 
{\ours} and {\oursb} model reactions through a two-step process 
of reaction center identification and synthon completion, allowing for the 
interpretability of reaction centers in the predicted reactants.}
%
Overall, \ours and \oursb achieve even better performance than the 
methods with full coverage, measured on the entire test set.}
%
%

%

\subsubsection*{Comparison with template-based (\tb) methods}
\label{sec:results:overall:temp}

\ours and \oursb achieve competitive performance with that from the \tb methods. 
With reaction types known, \ours achieves either the second or the third on all the top 
accuracies; with reaction types unknown,
\oursb achieves the best performance on top-1 (54.1\%), and either the second or the third
on all the other top accuracies.
For example, with reaction types unknown, \oursb is the second best on top-3 accuracy, with 
3.8\% difference from the best performance of \localretro~{\cite{chen2019localretro}}; 
\oursb slightly underperforms the second-best baseline \mhnreact~{\cite{seidl2021modern}} on top-10 (86.7\% compared to 87.9\% from \mhnreact), but outperforms 
\mhnreact on all the other metrics.
\localretro is a very strong \tb method. It extracted 731 templates from the benchmark training data, 
whereas other \tb methods have much more templates (11,647 for \gln and 9,162 for \mhnreact).
Therefore, \localretro could achieve better template selection over a small template set compared to others over much larger template sets. 
However, \localretro may suffer from scalability issues on 
large datasets because it scores all the reaction templates on all the potential reaction centers (i.e., all atoms and all bonds) 
in the product molecules. 
In general, all \tb methods may not generalize well to 
reactions that are not covered by the templates~\cite{somnath2021learning}. 
{
In terms of coverage on the test set, Table~\ref{tbl:overall} shows that the templates used in {\retrosim}, 
{\neuralsim} and {\mhnreact} can cover the entire test set, 
while the templates used in {\gln}~{{\cite{dai2019}}} and {\localretro} cannot (93.3\% for {\gln} and 98.1\% for {\localretro}).}
Unlike 
{{\tb} methods}, \ours does not use reaction templates, and only scores all the bonds and atoms once for reaction 
center identification, and thus is much more scalable in inference. It learns the patterns from training data and thus 
has a better chance to discover new patterns from the training data that are not covered by templates. 
%

\subsection*{Individual Module Performance}
\label{sec:results:module}
%


\begin{table*}[!b]
	\captionof{table}{\textbf{Module performance comparison {on reaction center identification and synthon completion} in top-$k$ accuracy (\%)}}
	\label{tbl:module}
  \vspace{-8pt}    
\begin{center}
\begin{small}
\begin{threeparttable}
	\begin{tabular}{
			@{\hspace{0pt}}l@{\hspace{3pt}}
			@{\hspace{3pt}}l@{\hspace{3pt}}
			@{\hspace{-10pt}}r@{\hspace{3pt}}
			@{\hspace{3pt}}c@{\hspace{3pt}}
			@{\hspace{3pt}}c@{\hspace{3pt}}
			@{\hspace{3pt}}c@{\hspace{3pt}}
			@{\hspace{3pt}}c@{\hspace{3pt}}
			@{\hspace{3pt}}r@{\hspace{3pt}}
			@{\hspace{3pt}}c@{\hspace{3pt}}
			@{\hspace{3pt}}c@{\hspace{3pt}}
			@{\hspace{3pt}}c@{\hspace{3pt}}
			@{\hspace{3pt}}c@{\hspace{0pt}}
		}
			\toprule
			\multirow{2}{*}{Module} & \multirow{2}{*}{Method} & \multirow{2}{*}{\parbox{0.07\textwidth}{Coverage (\%)}} & \multicolumn{4}{c}{Reaction type known} & & \multicolumn{4}{c}{Reaction type unknown}\\
			\cmidrule(lr){4-7} \cmidrule(lr){9-12}
			&  &  & 1 & 2 & 3 & 5 & &  1 & 2 & 3 & 5 \\	
			\midrule
			\multirow{4}{*}{\parbox{0.08\textwidth}{Reaction center identification}}
			 & \gtog  & 97.9 & 90.2 (92.1) & 94.5 (96.5) & 94.9 (96.9) & 95.0 (97.0) & & 75.8 (77.4) & 83.9 (85.7) & 85.3 (87.1) & 85.6 (87.4) \\
			 & \graphretro    & 95.0 & 84.6 (89.1) & 92.2 (97.1) & 93.7 (98.6) & 94.5 (99.5) & & 70.8 (74.5) & 85.1 (89.6) & 89.5 (94.2) & 92.7 (97.6) \\
			 & \retroprime    & 100.0 & 84.6 (84.6) & 94.0 (94.0) & 96.7 (96.7) & 97.9 (97.9) & & 65.6 (65.6) & 81.3 (81.3) & 87.7 (87.7) & 92.0 (92.0) \\
			 &  \ours & 97.5 & 84.3 (86.5) & 94.6 (97.0) & 96.5 (99.0) & 97.0 (99.5) & & 69.5 (71.3) & 85.6 (87.8) & 90.8 (93.1) & 94.8 (97.2) \\   
			 &  \oursb & 97.5 & 85.0 (87.2) & 94.1 (96.5) & 96.2 (98.7) & 97.3 (99.8) & & 69.3 (71.1) & 85.4 (87.6) & 91.1 (93.4) & 94.7 (97.1) \\
			 \midrule
			\multirow{4}{*}{\parbox{0.08\textwidth}{Synthon completion}}
			 & \gtog  &  100.0 &  66.8 & - & 87.2 & 91.5 & & 61.1 & - & 81.5 & 86.7 \\
			 & \graphretro    & 99.7 &  77.4 (77.6) & 89.5 (89.8) & 94.2 (94.5) & 97.6 (97.9) & & 75.6 (75.8) & 87.4 (87.7) & 92.5 (92.8) & 96.1 (96.4) \\
			 & \retroprime    & 100.0 &  75.0 &  -   & 88.9 & 90.6 & & 73.4 &  -   & 87.9 & 89.8 \\
			 & \ours          & 100.0 &  72.8 & 85.6 & 90.2 & 93.0 & & 73.3 & 84.6 & 89.6 & 92.8 \\
			\bottomrule
	\end{tabular}%
		\begin{tablenotes}[normal,flushleft]
		\begin{footnotesize}

	\item 
	{Columns with 1, 3, 5 and 10 present top-1, top-3, top-5 and top-10 accuracies, respectively.
	Column ``Coverage(\%)'' represents the percentage of test reactions that the modules of methods can be applied to.}
	``($\cdot$)'': the accuracy within the covered reactions.
	{All the baseline results are reported in their original papers, where ``-'' represents that the corresponding results are not reported.}

	\par 
		\end{footnotesize}
	\end{tablenotes}
\end{threeparttable}
\end{small}
  \vspace{-10pt}    
\end{center}
\end{table*}

Following the typical evaluation for \semitb methods as in literature~\cite{somnath2021learning}, Table~\ref{tbl:module}
presents the {individual} performance of the two modules - reaction center identification and synthon completion 
in \semitb methods. 
%
{In Table~{\ref{tbl:module}}, for the reaction center identification module, 
the top-$k$ accuracy measures the ratio of test products 
that have the ground-truth reaction center correctly predicted among the top-$k$ predictions. 
In the synthon completion module, 
the synthon completion is done according to the \emph{ground-truth} reaction center, not the \emph{predicted}
reaction center; 
the top-$k$ accuracy measures the ratio of test products that 
have the ground-truth reactants correctly predicted among the top-$k$ predictions.
Please note that here ``ground-truth" reaction center means the reaction center as appears in the benchmark data per our reaction 
center definition.}


\subsubsection{Comparison on reaction center identification}
\label{sec:results:module:center}

Among all the \semitb methods, the definitions of reaction centers vary.
In \gtog, reaction centers are referred to as the only one newly formed bond during the reaction, 
and reaction center identification predicts whether there is such a new bond (and its location) or not in the products
as in a classification problem. This reaction center definition and classification can cover 97.9\% of the test data (the rest 2.1\% 
correspond to multiple newly formed bonds). 
{\graphretro} defines the reaction center as the newly formed bond (\bondfmC as defined in Section ``Reaction Centers 
with New Bond Formation''
but without induced bond changes), 
the changed bond (\bondcgC as in ``Reaction Centers with Bond Type Change'') and the single atom with
changed hydrogen count (\atomC as in ``Reaction Centers with Single Atoms''), which in total covers 95.0\% of the reactions in the test set.
{\retroprime} aims to identify all the atoms involved in the reactions as reaction centers, which covers all the reactions in the test set.
%
{\ours} extends the definition of the reaction center in {\graphretro} with induced bond type change 
and atom charge changes, covering 97.5\% of the test set.

Due to the data leakage issue as revealed by Yan \etal~\cite{yan2020retroxpert} (i.e., reaction center is given in both the training and test data), 
the reported \gtog reaction center identification performance as cited in Table~{\ref{tbl:module}} 
is overestimated, {but the updated results have not been provided in their Github.} 
\graphretro uses two functions, one for bonds and one for atoms, to predict reaction centers. While these functions are able to 
predict well when such bonds and atoms are truly reaction centers (i.e., performance in parentheses in Table~\ref{tbl:module}), 
\graphretro's reaction center definition covers the least (95\%) of the test set compared to the other methods, resulting 
in still low accuracies (i.e., performance outside parentheses) over the test set. 
\retroprime has a very generic definition of reaction centers -- any atoms involved in the reactions, and uses one unified model 
to predict these atoms. However, as these atoms may experience different changes (e.g., connected to or disconnected from other atoms), 
a unified model not customized to specific changes may not suffice, leading to overall relatively low accuracies compared to other methods, 
particularly when reaction types are unknown. 
\ours and \oursb have the most comprehensive definition of reaction centers (Section ``Reaction Center Identification'') 
with high coverage (97.5\%) on the test set. 
In addition, \ours and \oursb use a specific predictor for each of the reaction center types. Therefore, they achieve the best overall accuracy 
among the entire test set, as well as good performance over the reactions covered by its reaction center definition. 
%

\subsubsection{Comparison on synthon completion}
\label{sec:results:module:synthon}

To compare synthon completion performance, all the {ground-truth} reaction centers defined by different methods are given and used to 
start the completion processes.
\gtog predicts only bond establishment in its reaction center identification and thus has to deal with any associated changes such as 
bond type change in its synthon completion process, which complicates the synthon completion prediction. 
Therefore, its performance on synthon completion is the worst among all the methods. 

\graphretro formulates the synthon completion as a classification problem over all the subgraphs that can realize the difference 
between the synthons and reactants. 
Therefore, its synthon completion is not guaranteed to work for all possible 
products (e.g., 99.7\% coverage over the test set), particularly if the needed subgraph is not included in the pre-defined vocabulary. 
Among all the products that \graphretro can handle, its synthon completion performance is the best, due to that classification 
can be much easier than generation as all the other methods do. 
However, since \graphretro does not do well in reaction center identification, overall, 
it does not outperform other methods in retrosynthesis prediction as Table~\ref{tbl:overall} demonstrates.
{In addition, the synthon completion module of {\graphretro} may fail to accurately estimate the likelihoods of leaving groups,
due to the ignorance of overall structures of predicted reactants.
Such inaccurate likelihood estimation may aggravate the bias of beam search
and reduce the diversity of predicted reactants as discussed in \graphretro~\cite{somnath2021learning}. 
}

{\retroprime} transforms the synthons to reactants using a Transformer, but similarly to {\gtog}, 
also needs to deal with additional predictions such as bond type change.
\retroprime's synthon completion performs reasonably well on top-1 accuracies. 
Together with its good top-1 accuracy on reaction center identification, \retroprime achieves the best top-1 accuracy
with reaction type known as demonstrated in Table~\ref{tbl:overall}. 
{{\retroprime} uses a rule 
to enumerate predicted reactants from the top-3 reaction centers, limiting the potential diversity of predicted reactants.
On average, 
\retroprime underperforms \ours, particularly on top-3 and top-5 accuracies in synthon completion. 
}

{\ours} does not use {\brics} fragments in synthon completion because the fragment information is not available for the substructures 
that will be attached to synthons. 
Compared to \graphretro, \ours leverages a generative process to add substructures 
to synthons in synthon completion, which is inherently more difficult than classification as in \graphretro but could be 
generalizable to new products and reactants.
{Meanwhile, {\ours} does not limit the number of reaction centers within the top-10 predicted reactants,
and thus increases the diversity of predicted reactants.}

{Although \ours does not outperform \graphretro in the synthon completion module alone, its generative process 
allows \ours to consider all the intermediate molecular structures and more accurately estimate the likelihood 
of each completion action, conditioned on the reaction centers and the corresponding synthons from its 
reaction center identification module (i.e., not the ground-truth reaction centers).
{Consequently, despite employing a beam search strategy similar to that of \graphretro, the generative process of \ours could alleviate the bias of beam search on most possible reaction centers by accurately estimating the likelihood of the completed reactants.}
{In contrast,} 
\graphretro may not generalize well, particularly given that \graphretro's reaction center identification does not perform well with respect 
to the ground-truth reaction centers (i.e., in the top panel of Table~\ref{tbl:module}), 
but its synthon completion module is trained using the ground-truth reaction centers 
(i.e., in the bottom panel of Table~\ref{tbl:module}). }
%

\subsection*{Performance on Different Reaction Types}
\label{sec:results:class}
%
%

\begin{table*}[!t]
	\captionof{table}{\textbf{\ours performance on different reaction types}}
	\label{tbl:class}
  \vspace{-8pt}    
\begin{center}
\begin{small}
\begin{threeparttable}
	\begin{tabular}{
			@{\hspace{2pt}}l@{\hspace{5pt}}
			@{\hspace{2pt}}r@{\hspace{5pt}}
			@{\hspace{5pt}}r@{\hspace{5pt}}
			@{\hspace{5pt}}c@{\hspace{5pt}}
		    	@{\hspace{5pt}}c@{\hspace{5pt}}
			@{\hspace{5pt}}c@{\hspace{5pt}}
			@{\hspace{5pt}}c@{\hspace{5pt}}
			@{\hspace{2pt}}c@{\hspace{2pt}}
			@{\hspace{5pt}}c@{\hspace{5pt}}
			@{\hspace{5pt}}c@{\hspace{5pt}}
			@{\hspace{5pt}}c@{\hspace{5pt}}
			@{\hspace{5pt}}c@{\hspace{5pt}}
		}
			\toprule
			 \multirow{2}{*}{Type Name} & \multirow{2}{*}{\parbox{0.05\textwidth}{Percentage (\%)}} &  & \multicolumn{4}{c}{Reaction type known} & & \multicolumn{4}{c}{Reaction type unknown}\\
			\cmidrule(lr){4-7} \cmidrule(lr){9-12}
			& & & 1 & 3 & 5 & 10 & &  1 & 3 & 5 & 10 \\
			\midrule 
			heteroatom alkylation and arylation    &  30.3 &&  62.3&  84.1&  90.2&  94.4 & & 56.1&  77.2&  84.4&  91.3 \\
	       	acylation and related processes        &  23.8 &&  76.1&  93.9&  96.7&  97.6 & & 67.0&  87.3&  92.3&  95.4 \\
	       	deprotections                          &  16.5 &&  58.3&  87.2&  91.5&  93.9 & & 51.8&  76.5&  82.7&  87.9 \\
			C-C bond formation                     &  11.3 &&  48.1&  68.1&  75.7&  82.4 & & 37.2&  56.6&  67.9&  75.7\\
			reductions                             &   9.2 &&  72.5&  87.9&  91.8&  95.0 & & 52.7&  69.8&  78.1&  84.6 \\
			functional group interconversion      &   3.7  &&  50.5&  69.0&  75.5&  81.0 & & 42.4&  52.7&  60.9&  67.9 \\
			heterocycle formation                  &  1.8  &&  -   &   -  &   -  &  -   & &  -   &   -  &   -  &  -  \\
			oxidations                             &   1.6 &&  86.6&  91.5&  92.7&  95.1 & & 62.2&  80.5&  85.4&  91.5 \\
			protections                            &  1.4  &&  85.3&  89.7&  89.7&  89.7 & & 48.5&  67.6&  85.3&  86.8 \\
			functional group addition             &   0.5  &&  95.7&  95.7&  95.7&  95.7 & & 78.3&  82.6&  87.0&  87.0 \\
			\bottomrule
	\end{tabular}%
	\begin{tablenotes}[normal,flushleft]
	\begin{footnotesize}
	\item {\!Columns with 1, 3, 5 and 10 present top-1, top-3, top-5 and top-10 accuracies, respectively. }
	{Column ``Percentage(\%)'' represents the percentage of reactions in the test set belonging to the specific reaction type.}
	{``-'' represents that the corresponding results are not available due to the lack of coverage.}
		\par 
		\end{footnotesize}
	\end{tablenotes}
	%
\end{threeparttable}
\end{small}
  \vspace{-10pt}    
\end{center}
\end{table*}

Table~\ref{tbl:class} presents the top-$k$ accuracy ($k$=1,3,5,10) of the reactions of different types.
This method appears to predict certain reaction types more accurately than others as shown in Table~\ref{tbl:class}. 
This is likely due to the relative structural diversity among potential reactants, particularly for substrates that can all provide the same products. 
For example, in the case of oxidations, only a very limited set of substrates can be utilized to generate a ketone, 
most commonly the oxidation of an alcohol, although ketones can certainly be accessed through other types of reactions as well.
This leads to the relatively higher accuracies of \ours on the reactions of oxidations (e.g., 62.2\% top-1 accuracy with 
reaction type unknown). 
In terms of reductions, however, numerous substrates could be utilized to generate an amine, 
including reductions of amides, nitro groups, and nitriles to name a few.
In addition, there are numerous methods to access the same amines through various structurally unique deprotection reactions. 
The number of methods available to access a specific functional group, therefore, may make it more difficult to 
accurately predict which method has been used for a specific molecule, leading to the lower accuracies
on reactions of deprotections
(e.g., 58.3\% top-1 accuracy with reaction type known). 
This would certainly be the case in carbon-carbon bond forming reactions as well, which can be assembled 
in a number of ways from various substrates, potentially leading to a somewhat lower prediction success rate 
(e.g., 37.2\% top-1 accuracy with reaction type unknown).
In addition, as shown in our case studies, in molecules containing more than one functional group, there are often 
multiple ways in which that molecule can be assembled by targeting each individual functional group as the reaction center.
This means that there are multiple valid reaction pathways which could be considered by synthetic chemists 
in order to most efficiently construct a molecule.
{Please note that \ours is designed to predict reactions that involve three types of reaction centers:
1) a single newly formed bond with induced changes in bond types; 
2) a single changed bond;
3) a single atom with a fragment removed.
As a result, \ours could not fully cover reaction types such as rearrangement, isomerization, cyclization and click reactions, which involve multiple changes in bond formation or atom detachment.
%
This illustrates \ours's limitation in handling all possible reaction types.
It is worth noting that other semi-template-based methods such as \gtog and \graphretro, also share this limitation.
Therefore, developing an effective semi-template-based method that overcomes this limitation could be an interesting future research direction.} 
\subsection*{{Performance of Ensemble-based Methods}}
\label{sec:results:ensemble}
%

We also compared the performance of an ensemble of \ours, referred to as \oursens, 
with \augtrans~\cite{Tetko_2020} and \rsmiles~\cite{Zhong2022}, both of which test each target product multiple times
and are strong baselines. 
\augtrans and \rsmiles represent each target molecule using multiple non-canonical but equivalent SMILES strings, and use 
the multiple SMILES strings during model training and testing.
By combining the predictions from the multiple SMILES strings of the same target product, 
these methods have the choice to explore a larger reaction subspace seeded by the SMILES strings, and thus achieve better 
prediction performance. 
Compared to the SMILES strings, \ours uses molecular graph representations, and thus each molecule can only have a unique 
representation. Instead of augmenting molecule representations but still being able to explore a larger reaction subspace as 
\augtrans and \rsmiles do, \oursens tests each molecule multiple times using multiple {\ours} models. 
Details of \oursens are available in the supplementary Note 2. 

\begin{table*}[t]
	\captionof{table}{\textbf{{Overall comparison on retrosynthesis prediction {between {\oursens} and baselines }with test set augmentation in top-$k$ accuracy (\%)}}}
	\label{tbl:overall_aug}
  \vspace{-8pt}    
\begin{center}
\begin{small}
\begin{threeparttable}
	\begin{tabular}{
			@{\hspace{2pt}}l@{\hspace{5pt}}
			@{\hspace{2pt}}l@{\hspace{5pt}}
			@{\hspace{2pt}}l@{\hspace{5pt}}
			@{\hspace{5pt}}c@{\hspace{5pt}}
			@{\hspace{5pt}}c@{\hspace{5pt}}
			@{\hspace{5pt}}c@{\hspace{5pt}}
			@{\hspace{5pt}}c@{\hspace{5pt}}
		}
			\toprule
			\multirow{2}{*}{Dataset} &  \multirow{2}{*}{\parbox{0.1\textwidth}{Method type}} & \multirow{2}{*}{Method} & \multicolumn{4}{c}{Reaction type unknown}\\
			\cmidrule(lr){4-7} 
			& & & 1 & 3 & 5 & 10 \\	
			\midrule 
			\multirow{3}{*}{All reactions} 
			& \multirow{2}{*}{\parbox{0.1\textwidth}{{\tf}}}
			 & \augtrans~\cite{Tetko_2020} & 52.7 &   73.4  & 79.1 & 83.7 \\
			 & & \rsmiles~\cite{Zhong2022} & \textbf{56.5} & \textbf{79.4} & \textbf{86.0} & \textbf{91.0} \\
			 \cmidrule{2-7}
			 & {\parbox{0.15\textwidth}{{\semitb}}}
			 & \oursens & 56.4 & 78.8 & 85.2 & 90.5 \\
			 \midrule
			 \multirow{3}{*}{\parbox{0.25\textwidth}{Reactions covered by \ours}} 
			& \multirow{2}{*}{\parbox{0.1\textwidth}{{\tf}}}
			& \augtrans~\cite{Tetko_2020} & 54.1 &   75.5  & 81.4 & 85.8 \\
			& & \rsmiles~\cite{Zhong2022} & 56.8 & 79.7 & 86.2 & 91.3 \\
			\cmidrule{2-7}
			 & {\parbox{0.15\textwidth}{{\semitb}}}
			 & \oursens & \textbf{57.8} & \textbf{80.7} & \textbf{87.3} & \textbf{92.7} \\
			\bottomrule
	\end{tabular}%
		\begin{tablenotes}[normal,flushleft]
		\begin{footnotesize}
	\item \!Columns with 1, 3, 5 and 10 present top-1, top-3, top-5 and top-10 accuracies, respectively. 
	Best top-$k$ accuracy values among the methods of each type are in \textbf{bold}.
		\par 
		\end{footnotesize}
	\end{tablenotes}
\end{threeparttable}
\end{small}
  \vspace{-10pt}    
\end{center}
\end{table*}

Table~{\ref{tbl:overall_aug}} presents the comparison among \oursens, \augtrans and \rsmiles on
top-$k$ accuracy ($k$=1,3,5,10) over all the reactions and the reactions covered by {\ours}, both with the reaction type unknown.
{Please note that the performance of {\augtrans} and {\rsmiles} 
on reactions with known types is not available in the respective papers~\cite{Tetko_2020,Zhong2022}, and the methods also cannot be easily extended to handle known reaction types.} 
In Table~{\ref{tbl:overall_aug}}, all the methods test each molecule 20 times, that is, \augtrans and \rsmiles augment each target molecule with 20 SMILES strings, 
and \oursens uses an ensemble of 20 models to test each molecule.
The results of \augtrans and \rsmiles are calculated using the source code and data available from the respective papers. 
Table~{\ref{tbl:overall_aug}} shows that \oursens achieves competitive performance with the best baseline \rsmiles.
Over all the reactions, {\oursens} achieves almost the best performance on top-1 (56.4\%, compared to 56.5\% for {\rsmiles}), 
and only slightly underperforms the best baseline {\rsmiles} on top-3, top-5 and top 10 (78.8\% vs 79.2\% 
on top-3; 85.2\% vs 86.2\% on top-5; 90.5\% vs 91.0\% on top-10).
Over the reactions covered by {\ours}, {\oursens} outperforms the baseline {\rsmiles} on top-1 accuracy at 1.76\%, on top-3 accuracy at 1.25\%, 
on top-5 accuracy at 1.28\%, and on top-10 accuracy at 1.53\%.
Compared to {\rsmiles}, which is an end-to-end black-box that directly transfers product SMILES string to reactant SMILES strings, 
\ours provides certain interpretability of the predicted reaction centers, and what reactants are generated from them. 
More details about the comparison on different reaction types and 
on reactions covered by {\oursens} are available in the supplementary Note 2.
%
 

\subsection*{Case Study}
\label{sec:results:case}
%

\begin{figure*}[!t]
    \centering
    \includegraphics[width=0.95\textwidth]{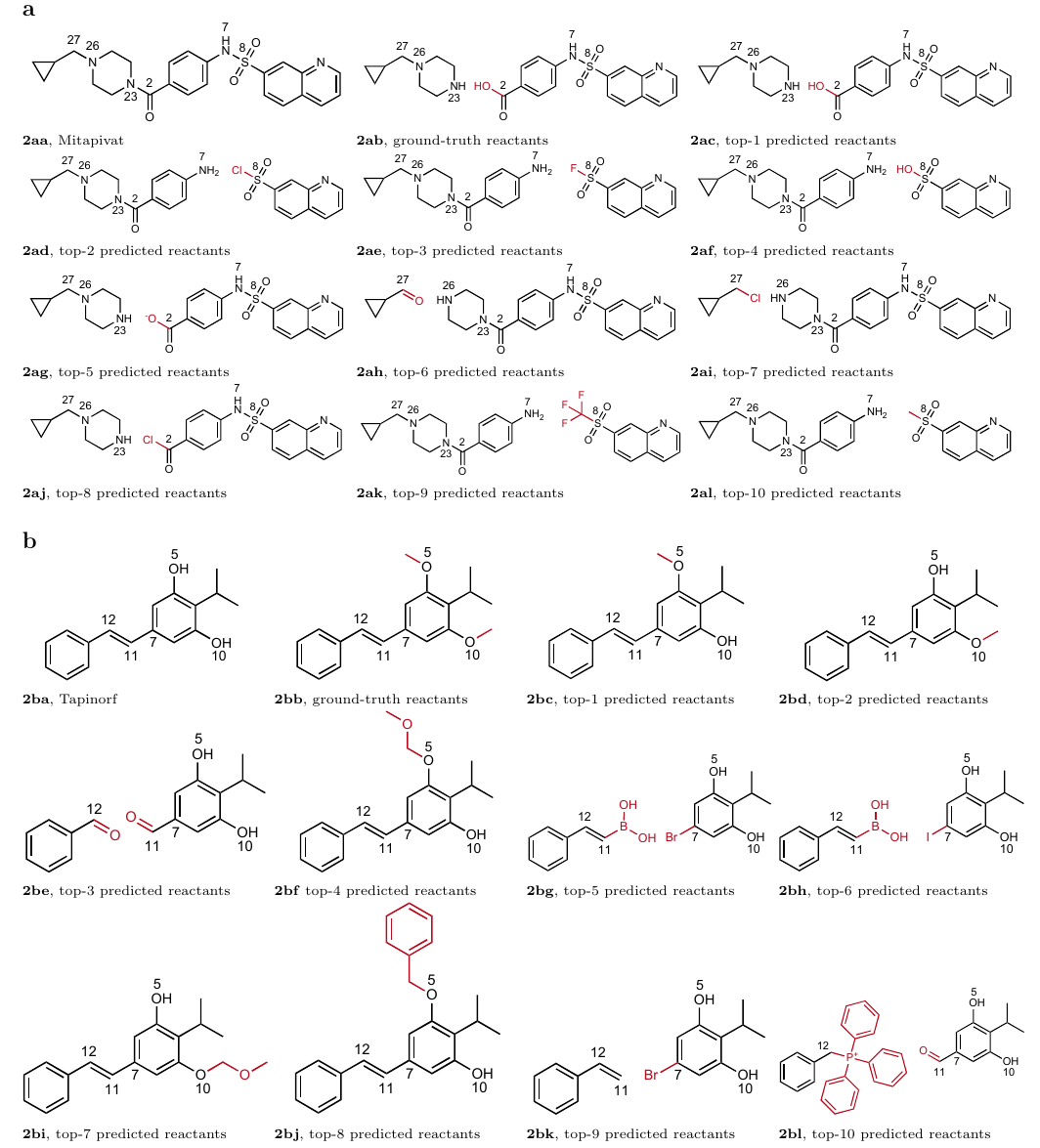}
    \caption{\textbf{Predicted reactions by \ours for two newly approved drug molecules.}
                 \textbf{a} Predicted reactions by \ours for Mitapivat;
                 \textbf{b} Predicted reactions by \ours for Tapinorf.
		Numbers next to each atom are the indices of the atoms. 
		Atoms with same indices in different subfigures are corresponding to each other. 
		{Atoms and bonds colored in red are leaving groups for synthon completion.
		Molecules with labels ending in \textbf{a} are product/target molecules; 
		molecules with labels ending in \textbf{b} are the reactants reported in patents; 
		molecules with labels ending in \textbf{c-l} are the top predicted reactants.}}
    \label{fig:fig2}
\end{figure*}

\ours can predict multiple reactions for each product due to multiple predicted reaction centers.
This variability could be useful for chemical synthesis in order to consider all possible reaction strategies. 
In order to illustrate the predictive power of \ours, we have highlighted {the top-10 predicted reactants by \ours with reaction types unknown
for four newly approved drug molecules in 2022, including Mitapivat, Tapinorf,
Mavacamten, and Oteseconazole~\cite{fda_newdrugs}. 
Among them, the predicted reactants for Mitapivat and Tapinorf are presented in Figure~\textbf{2aa} 
and \textbf{2ba} 
which will be discussed later; the results and the discussions for Mavacamten and Oteseconazole are available in 
{Supplementary Figure 1, Supplementary Figure 2 and }Supplementary Note 3. 
Note that these drugs are not included in our training, validation, or testing data. 
Therefore, how \ours works on these drugs truly indicates its predictive power for new molecules.}

{Mitapivat as in Figure~\textbf{2aa} 
is a drug approved for hereditary hemolytic anemias in 2022~\cite{Al_Samkari_2021}. 
%
%
{The synthetic route within the patent~{\cite{mitapivat}} reporting the discovery of Mitapivat utilizes an amide coupling reaction to form the C2-N23 bond (Figure~\textbf{2ab}). 
This is correctly predicted by {\ours} as the top-1 reaction (Figure~\textbf{2ac}). 
As indicated by the top-5 reaction (Figure~\textbf{2ag}), 
{\ours} also predicts that the amide coupling reaction could be performed with the 
carboxylate salt of one of the reactants, a useful reactant under the right pH conditions.
\ours also predicts that the acyl chloride as the substrate in this transformation would also react with the amine group and produce the desired molecule (Figure~\textbf{2aj}), 
In addition, \ours identifies the N7-S8 bond of {sulfonamide linkage} as the reaction center (e.g., Figure~\textbf{2ad}, \textbf{2ae}, \textbf{2af}, \textbf{2ak}, \textbf{2al}). 
Most impressively, \ours predicts various S8 sulfonyl groups reacting with the N7 amine group, such as sulfonyl chloride (Figure~\textbf{2ad}), 
sulfonyl fluoride (Figure~\textbf{2ae}) 
and sulfonic acid (Figure~\textbf{2af}), 
which {are theoretically feasible for the formation of the N7-S8 bond}. 
\ours also predicts that the N26-C27 bond could be the reaction center and formed by the N26 amine group reacting {through a reductive amination with ketone in Figure~\textbf{2ah} 
or through a nucleophilic substitution with the chloride in Figure~\textbf{2ai}. 

{Tapinarof as in Figure~\textbf{2ba} 
is a drug approved for plaque psoriasis and atopic dermatits~\cite{Keam_2022_tapinarof}.
The reported synthesis in patent~\cite{tapinorf} constructs this drug by removing the protecting groups on O5 and O10 (Figure~\textbf{2bb}). 
{\ours} correctly predicts the deprotection of the methyl groups on O5 (Figure~\textbf{2bc}) 
or O10 (Figure~\textbf{2bd}), 
which would work to produce the desired molecule, although the ground truth failed to be predicted due to the limitation of reaction centers.
{Similarly, {\ours} generates possible reactants that contain different types of protected alcohols, as seen with the methoxymethyl groups on O5 and O10 in Figure~\textbf{2bf} and Figure~\textbf{2bi} and the benzyl-protected O5 in Figure~\textbf{2bj}.
%
Most impressively, \ours also identifies the alkene linkage between C11 and C12 (Figure~\textbf{2be} and \textbf{2bl})
and the C-C bond between C7 and C11 (Figure~\textbf{2bg}, \textbf{2bh}, ad \textbf{2bk}) 
as reaction centers with various coupling reactions.
These coupling reactions include McMurry coupling~\cite{Duan2006} (Figure~\textbf{2be}), 
Wittig coupling~\cite{Robiette2006} (Figure~\textbf{2bl}) 
and Suzuki coupling~\cite{Miyaura1995} (Figure~\textbf{2bg} and \textbf{2bh}). 
}

\begin{figure*}[!h]
    \centering
    \includegraphics[width=0.95\textwidth]{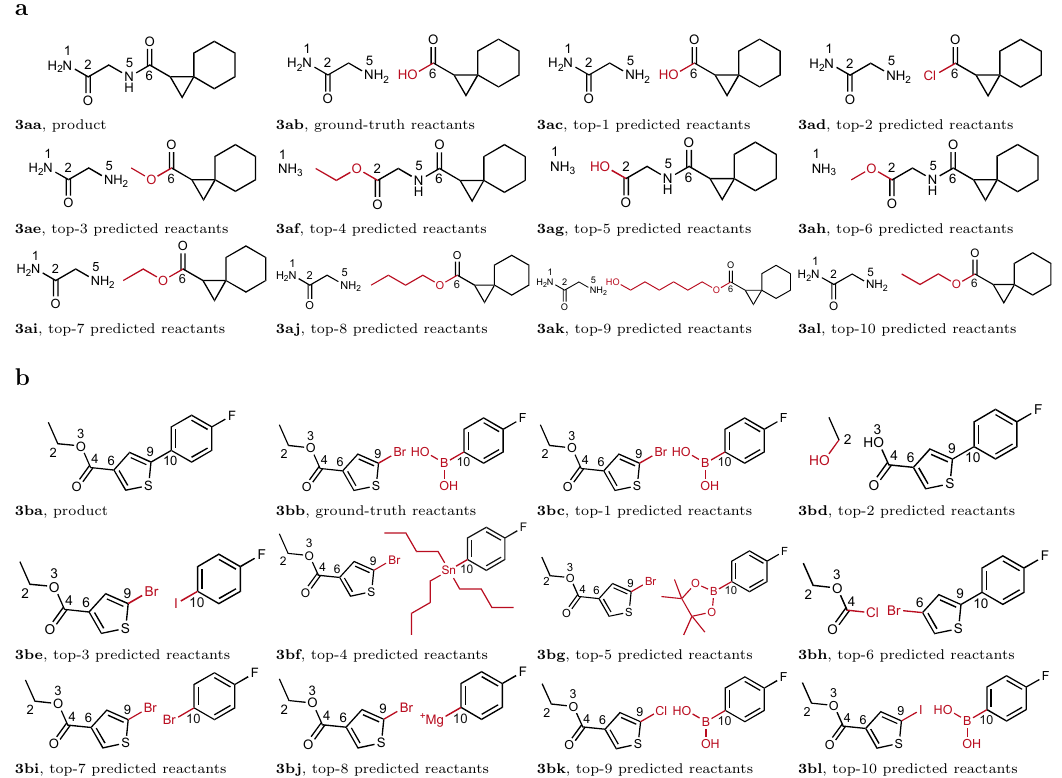}
    \caption{\textbf{Predicted reactions by \ours for two test molecules in USPTO-50K.}
                 \textbf{a} Predicted reactions by \ours for product ``NC(=O)CNC(=O)C1CC12CCCCC2''; 
                 \textbf{b} Predicted reactions by \ours for product ``CCOC(=O)c1csc(-c2ccc(F)cc2)c1''.
		Numbers next to each atom are the indices of the atoms.
		Atoms with same indices in different subfigures are corresponding to each other.
		{Atoms and bonds colored in red are leaving groups for synthon completion.
		Molecules with labels ending in \textbf{a} are product/target molecules; 
		molecules with labels ending in \textbf{b} are the ground-truth reactants in USPTO-50K; 
		molecules with labels ending in \textbf{c-l} are the top predicted reactants.}}
    \label{fig:fig3}
\end{figure*}

In addition, we also highlighted two molecules in the test set and their predicted reactions by \ours with reaction types unknown in Figure~{3a} and {3b}, respectively. 
The product in Figure~\textbf{3aa} 
contains amide linkages and was assembled in the patent literature utilizing amide coupling reactions (ground truth in Figure~\textbf{3ab}). 
\ours correctly predicted this coupling as the top-1 reaction for the construction of this molecule (Figure~\textbf{3ac}). 
The other reactions predicted, however, are also very instructive into the strengths and limitations of \ours. 
In Figure~\textbf{3aa}, 
the product has two amide groups in the side chain of the molecule. 
\ours identified both of these linkages as potential reaction centers (e.g., in Figure~\textbf{3ac} 
between N5 and C6; in Figure~\textbf{3ag} 
between N1 and C2).
Typically, chemists would disconnect the molecule at the C6 amide carbonyl 
rather than C2 so that a fully elaborated side chain can be introduced to 
complete the molecule. 
This approach would generally be considered more efficient since its reaction introduces more complexity into the molecule in a single step and would therefore be predicted to limit the total number of steps necessary to construct the molecule. 
In some limited cases, however, it may be necessary to introduce the nitrogen at N1 last (e.g., in Figure~\textbf{3af}-\textbf{3ah}), 
so this should also be considered a feasible reaction. 
In addition to the typical amide coupling strategy, which takes place between an amine and a carboxylic acid, \ours also correctly identifies the reaction of the amine with an acid chloride to make the same bond (Figure~\textbf{3ad}). 
Although this was not the strategy utilized in the ground-truth study, this strategy would certainly be expected to work in this case for construction of this molecule. 
The other common reaction that was predicted for this example was the nucleophilic addition of the N5 (or N1) amine into the C6 (or C2) carbonyl of an ester (N5-C6 - Figure~\textbf{3ae}, \textbf{3ai}, \textbf{3aj}, \textbf{3ak}, \textbf{3al} and N1-C2 - Figure~\textbf{3ag} and \textbf{3ah}).
%
This type of reaction, which is essentially a transamidation reaction, should also work to provide the product. 
Interestingly, however, \ours predicts several different esters as substrates for this transformation 
(Figure~\textbf{3ac}, \textbf{3ae}, \textbf{3ai}, \textbf{3aj}, \textbf{3ak} and \textbf{3al}).
%
While these are different substrates, the variation of the ester side chain in these cases would not typically be considered as greatly 
different by a synthetic chemist unless steric or electronic contributions affect the reactivity/electrophilicity of the ester carbonyl.

Retrosynthesis of the product in Figure~3b 
involves a C-C bond forming reaction between C9 and C10 (Figure~\textbf{3ba}). 
The disconnection of the carbon-carbon bond between the two aromatic rings, a heteroaromatic thiophene and a benzene ring in this case, represents the most obvious disconnection in the molecule. 
In this case, the top-1 reaction (Figure~\textbf{3bc}) 
predicted by \ours for this transformation is a Suzuki coupling~\cite{Miyaura1995}, 
a common metal-mediated coupling between a boronic acid reactant and a corresponding aryl halide. 
This common transformation is the same reaction observed in the ground truth (Figure~\textbf{3bb}). 
Interestingly, \ours also identifies additional permutations of this Suzuki reaction through 
changing the nature of the aryl halide (Figure~\textbf{3bk} and \textbf{3bl}). 
Traditionally, aryl chlorides (Figure~\textbf{3bk}) 
are less reactive than aryl bromides or iodides (Figure~\textbf{3bc} and \textbf{3bl}) 
for coupling reactions and in the past were considered unreactive in these reactions.
Newer methods~\cite{LeBlond_2001} 
using specially designed ligands, however, have made the use of such chlorides possible.
The other difference observed in the predicted Suzuki couplings is the use of a boronic ester (Figure~\textbf{3bg}) 
versus a boronic acid (Figure~\textbf{3bc}). 
Both boronic acids and boronic esters are common reagents for these transformations, with many being readily available from commercial sources. 
\ours also predicts that an esterification reaction at the C4 carboxylic acid would also work to produce the desired molecule
(Figure~\textbf{3bd}). 
While this is potentially not as synthetically useful for building the molecule, it is a reasonable transformation.
Most impressively, \ours also predicts other coupling reactions~\cite{Yin2006}  for the biaryl coupling reaction. 
These other methods include an Ullmann-type coupling~\cite{FANTA1974} 
(Figure~\textbf{3be} and \textbf{3bi}) 
a Stille coupling~\cite{Stille1986} (Figure~\textbf{3bf}), 
and a Kumada coupling~\cite{Tamao1972,LeBlond_2001} (Figure~\textbf{3bj}). 
This versatility predicted in the top-10 reactions may be of synthetic value for substrates if specific coupling methods fail or if the functionality necessary for one type of coupling reaction is not able to be easily prepared.

The above examples indicate that the predicted reactions from {\ours} rather than the ground truth could be 
still possible and synthetically useful.
Therefore, a more comprehensive evaluation strategy is needed not to miss those possible and potentially novel 
synthesis reactions. 
%

\subsection*{Diversity on predicted reactions}

\begin{figure*}[!h]
	\centering
	\includegraphics[width=0.9\linewidth]{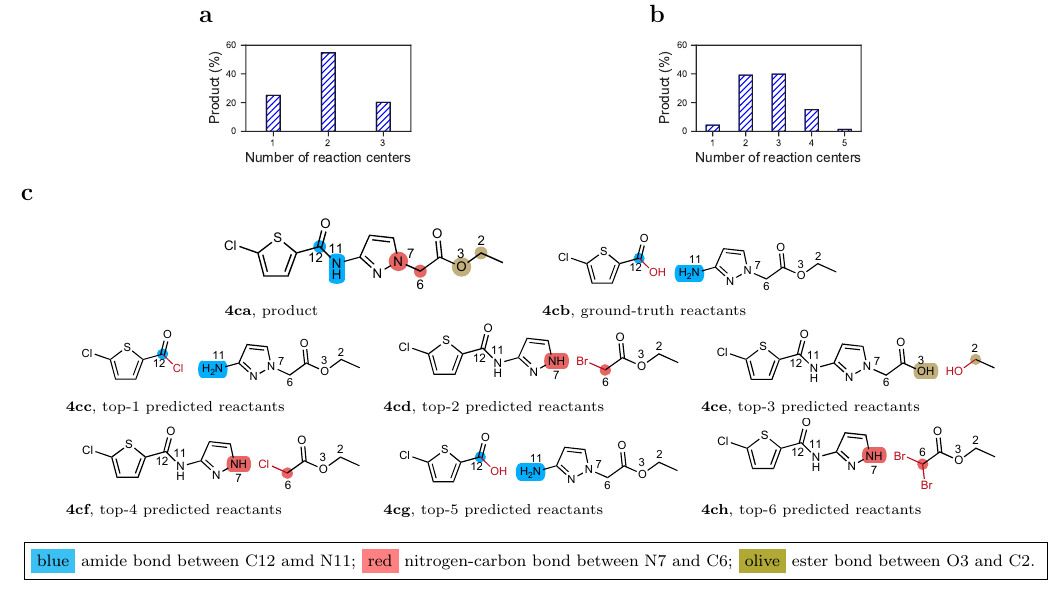}
	\label{fig:fig4}
	\caption{\textbf{Reaction center analysis in predicted reactions and a representative example.}
	\textbf{a} Percentage of products (Product (\%)) with the different number of predicted reaction centers and the third predicted reaction as the ground-truth reaction (i.e., hits at 3);
	\textbf{b} Percentage of products (Product (\%)) with the different number of predicted reaction centers and the fifth predicted reaction as the ground-truth reaction (i.e., hits at 5);
	\textbf{c} Predicted reactions by \ours for product ``CCOC(=O)Cn1ccc(NC(=O)c2ccc(Cl)s2)n''.
	Numbers next to each atom are the indices of the atoms. Atoms with the same indices 
	in different subfigures correspond to each other.
	Different reaction centers are highlighted in different colors (blue, red and olive).
	Atoms and bonds colored in red are leaving groups for synthon completion.
	Molecules with labels ending in \textbf{a} are product/target molecules; 
	molecules with labels ending in \textbf{b} are the ground-truth reactants in USPTO-50K; 
	molecules with labels ending in \textbf{c-h} are the top predicted reactants.
	}
\end{figure*}

Diversity in predicted reactions is always desired, as it has the potential to {enable the exploration of multiple synthesis routes.}
\ours has the mechanisms to facilitate diverse predictions: 
{The beam search strategy in \ours allows 
multiple reaction centers and multiple different attachments, and therefore potentially different scaffolds
and structures in the predicted reactants.}

To analyze the diversity of {\ours} results, we analyzed the reaction centers among the top-predicted reactions.
We identified a set of products such that their third or fifth 
predicted reactions are the ground truth, referred to as having a hit at 3 or 5, respectively. 
Please note each predicted reaction was scored using the sum of the log-likelihoods of all the predictions along the transformation 
paths from the product to its reactants (please refer to Section ``Inference''), 
and then ranked based on the score. 
Thus, the predicted reactions ranked above the ground truth have a higher likelihood than the 
ground truth. 
Given that \ours has demonstrated strong performance as in Table~{\ref{tbl:overall}} in scoring and prioritizing the ground-truth
reactions,  we assume that its likelihood calculation is reliable 
and therefore, the reactions ranked above the ground truth
might also be likely to occur.

Figure~{4a} and {4b} presents the distribution of products with hits at 3 or 5  over the number of reaction centers among 
predicted reactions ranked above 
the ground truth.
Figure~{4a} 
shows that more than 50\% of the products with a hit at 3 have their top-3 reactions from two different 
reaction centers; about 20\% of the products have their top-3 reactions from three different reaction centers. 
Figure~{4b} 
shows that for products with a hit at 5, almost 40\% have two reaction centers, and another 40\% have three reaction centers, 
among their top-5 predicted reactions; more than 10\% have four reaction centers. 
Thus, Figure~{4a} and {4b} 
clearly demonstrate that {the top predicted reactions were diverse, demonstrated by the different reaction centers they were derived from. 
Meanwhile, we acknowledge that the diverse, top predictions may still be errors
and thus, more reliable {wet-lab} 
experimental validation is needed. 
%
}

Figure~{4c} 
presents an example of very diverse reactions with diverse reaction centers predicted by {\ours}. 
For the product in Figure~\textbf{4ca}, 
{\ours} predicts three different reaction centers:
an amide bond (between C12 and N11), 
a nitrogen-carbon bond (between N7 and C6) and ester (between O3 and C2). 
%
The patent reported that the target molecule was synthesized from a carboxylic acid derivative and an amine using amide coupling with a widely-used coupling reagent, EDC (Figure~\textbf{4cb}). 
{\ours} predicted an acyl chloride-amine reactant pair as the top-1 result (Figure~\textbf{4cc}), 
a potentially viable and even high yielding synthetic approach. 
It also predicts three reactant pairs from the other two reaction centers as possible routes within the top 4
(Figure~\textbf{4cd} and \textbf{4cf} 
at which involve alkylation reactions to form the C6-N7 bond;
Figure~\textbf{4ce} 
at which forms the ester linkage between O3 and C2).

We also analyzed the reaction diversity by comparing the number of reaction centers in products with 
high reaction diversity and low reaction diversity. 
%
For each product, the diversity of its predicted reactions 
is represented by the distribution of all pairwise similarities of its predicted reactions, 
that is, lower reaction similarities indicate
higher reaction diversity.
Please note that the reaction similarity is only applicable to two reactions that share the same product.
Therefore, the product is not considered in the similarity calculation.
Formally, for reaction $R_1$: 
$\mol_1+ \mol_2 \rightarrow \molp$ and reaction $R_2$: $\mol_3+ \mol_4 \rightarrow \molp$, the similarity 
between $R_1$ and $R_2$ was calculated as follows,  
\begin{equation}
\label{eqn:sim}
\text{sim}(R_1, R_2) = 
	\frac{1}{2}\max(\text{sim}_m(\mol_1, \mol_3) + \text{sim}_m(\mol_2, \mol_4), 
	\text{sim}_m(\mol_1, \mol_4) + \text{sim}_m(\mol_2, \mol_3)), 
\end{equation}
where $\text{sim}_m()$ is a similarity function over molecules, calculated using Tanimoto coefficient over 
2,048-bit Morgan fingerprints of the molecules. 
For reaction $R_1$: $\mol_1 \rightarrow \mol_p$ and reaction $R_2$: $\mol_2 + \mol_3 \rightarrow \mol_p$, the similarity between them was calculated as follows,
\begin{equation}
\text{sim}(R_1, R_2) = \text{sim}_m(\mol_1, \mol_{2}+\mol_{3}),
\end{equation}
where $\mol_{2}+\mol_{3}$ denotes the composite molecule consisting of two disconnected components $\mol_{2}$ and $\mol_{3}$.
For reaction $R_1$: $\mol_1 \rightarrow \mol_p$ and reaction $R_2$: $\mol_2 \rightarrow \mol_p$, the similarity between them was calculated as follows,
\begin{equation}
\text{sim}(R_1, R_2) = \text{sim}_m(\mol_1, \mol_{2}).
\end{equation}
We clustered the products according to their reaction similarity distributions using the K-means
clustering algorithm in Euclidean distances. The clustering algorithm is presented in Supplementary Algorithm 1 in Supplementary Note 4. 
Figure~{5a} presents the clustering results for products 
that have their ground-truth reaction correctly predicted among the top-10 predictions.
In Figure~{5a}, the first four clusters 
have on average lower reaction similarities (on average 0.46 among the four clusters; 0.41, 0.45, 0.45, 0.49 in each of the 
clusters, respectively), and thus are referred to as high-reaction-diversity clusters ({\HRD}); 
the other six clusters, referred to as low-reaction-diversity clusters ({\LRD}), have relatively 
higher reaction similarities (on average 0.58 for among the six clusters; 0.52, 0.53, 0.58, 0.62, 0.67, 0.67 in each of the 
clusters, respectively).

\begin{figure}[!t]
	\centering
	\includegraphics[width=\textwidth]{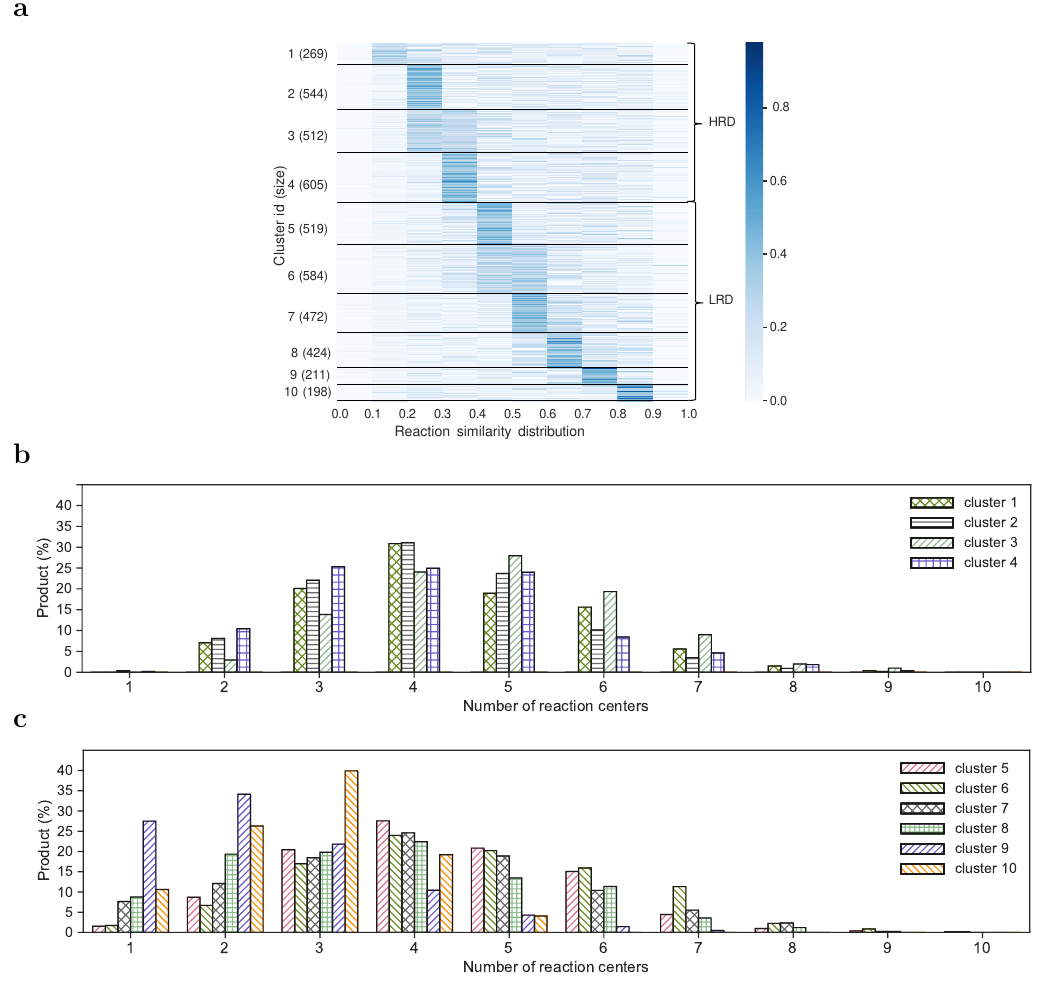}
	\label{fig:fig5}
	\caption{\textbf{Cluster analysis on test products based on similarities of their predicted reactions.}
	\textbf{a} Clustering on test products based on {similarities of their predicted reactions}. 
	The x-axis indicates the range of reaction similarities (e.g., the column between 0.1 and 0.2 indicates the range (0.1, 0.2]);     the y-axis shows the cluster ID and the cluster size. 
	Each row in the heatmap corresponds to the reaction similarity distribution of a product 
	belonging to a specific cluster; each block in the row corresponds to the frequency of reaction 
	similarities within each similarity range, and the block color represents the scale of the 
	frequency (e.g., a darker color indicates a higher frequency value).
	The clusters are labeled as `\HRD' for high-reaction-diversity clusters with average low reaction 
	similarities, and `\LRD' for low-reaction-diversity clusters with average high reaction similarities.
	\textbf{b} Test product distributions over the number of reaction centers of \HRD products.
	\textbf{c} Test product distributions over the number of reaction centers of \LRD products.}
\end{figure}

%
%
Figure~{5b} and {5c} 
present the distributions of the number of reaction centers 
in the products of these two clusters. 
Comparing Figure~{5b} and Figure~{5c}, 
{\HRD} {products} tend to have more reaction centers {in their predicted reactions} than those in {\LRD} {products}, 
and the number of reaction centers correlates well with reaction diversity (-0.8486 between the average {reaction} similarities 
and the number of reaction centers). 
Particularly,  the first cluster (in \HRD), which has the highest reaction diversity (lowest reaction similarity), 
has on average 4.41 reaction centers in the top-10 predicted reactions of each product,
compared to the average 3.92 reaction centers in the top-10 predicted reactions of each product in {\LRD} clusters. 
The ninth and tenth clusters, which have the lowest reaction diversity, 
have on average 2.57 reaction centers. 
These results clearly show the diversity of {\ours} predictions.

\section*{{Discussion}}
\label{sec:discussion}

\subsection*{Comparison among template-based, template-free and semi-template-based methods}

%
Template-based methods were first developed for retrosynthesis prediction. 
They match products into pre-defined templates that are extracted from training data or hand-crafted based on 
knowledge. 
A notable advantage of templates is that they can enable strong interpretability 
(e.g., each template may correspond to a certain reaction type, a chemical scaffold, or a reactivity pattern) and thus 
result in reactions that better conform to domain knowledge. 
They can also well fit the data if the templates are extracted from the data. 
However, they suffer from a lack of strong learning capabilities and a lack of generalizability, if 
the templates do not cover and cannot automatically discover novel reaction patterns. 
%

%
Template-free methods largely leverage the technological advancement in Natural Language Processing (NLP), 
including large-scale language models such as Transformer and BART~{\cite{Lewis_2020}}, and also many pre-training techniques. 
{Most of them} formulate a reaction as a SMILES string translation problem. 
Rather than enumerating pre-defined patterns (i.e., templates) as template-based methods do, template-free methods 
are equipped with much stronger learning capabilities from SMILES strings and can represent latent reaction transformation patterns in an 
operable manner. 
%
However, template-free methods sacrifice their
interpretability as it is non-retrieval to decipher why an atom (analogous to a token in NLP) is generated next along the SMILES strings, 
or what chemical knowledge the actions correspond to. 
In addition, as SMILES strings are a `flattened' representation of molecular graphs according to the atom orderings from a graph traversal,
template-free methods using SMILES strings only cannot fully leverage molecular structures, which ultimately determine molecule synthesizability and reaction types.
To mitigate this issue, some template-free methods either enrich the product SMILES representation with molecular graph information~\cite{Mao2021,seo2021gta,Wan2022}
 or
decode reactant SMILES strings from product molecular graphs~\cite{Tu2021}, which, however, require additional learning 
of the mapping from 
molecular graphs to SMILES and thus increase the learning complexity.
Semi-template-based methods, typically over molecular graphs, represent the most recent and also in general the best performing retrosynthesis prediction methods. 
They utilize the powerful graph representation learning paradigm to better capture molecule structures.
They also take advantage of graph (variational) auto-encoder frameworks or sequential predictions 
to empower the models with generative ability. 
More importantly, semi-template-based methods 
have the mechanism to enable diversity among predicted reactions, by allowing multiple samplings from the latent space.   
Meanwhile, semi-template-based methods {have two steps: (1) reaction center identification, and 
(2) synthon completion,}
better complying with how chemical reactions are 
understood and enabling certain interpretability of predicted reaction centers and derived reactants.
%
%
\ours is a semi-template-based method and achieves superior performance to other methods, demonstrating it as a state-of-the-art method
for retrosynthesis prediction.

\subsection*{{Comparison issues} among existing methods}

In our study of the baseline methods, several issues were identified 
among existing methods that make comparison across different methods hard. 
In Table~\ref{tbl:overall}, \retroxpert's results are from its updated GitHub~\cite{retroxpert_git}, as their  
results originally reported in their manuscript had a data leakage issue (all the reaction centers were implicitly given) 
and thus were overestimated~\cite{yan2020retroxpert}.
\gtog may also suffer from the data leakage issue as discussed in its github~\cite{gtogdisc}, 
but \gtog's results were only available from its original paper, though likely overestimated. 
In addition, 
there have been some reproducibility issues with G2G~{\cite{gtogresult}}
, as we also observed in our study.
All the methods except \neuralsim, \lvtrans, \dual and \retroformer published their code and datasets. 
Among these methods, most template-free methods including \scrop, \get, \chemformer,
\tiedformer, \gta and \augtrans used the same data split, which is, however, 
different from the benchmark data split used in the other methods.
For example, the training set of these template-free methods has 40,029 reactions, 
while the training set of the other methods including \ours has 40,008 reactions.
Even though all the methods 
adopted the same ratio (i.e., 80\%/10\%/10\% for training/validation/test set) to split the benchmark dataset, 
their splits, particularly their test sets, are not identical, making it hard to compare these methods.
In this manuscript, we adopted the data split used by the previous semi-template-based methods; 
for the template-free methods with different 
data splits, we still used the results reported by their authors.
We believe reproducibility and unbiased comparison (e.g., on the same benchmark data and same splits, generating 
the same amount of results to compare) among 
all the retrosynthesis prediction methods are critical to moving this research forward. They require dedicated
research,  implementation and regulatory effort from the entire research community, 
for example, by following the 
Open Science Policy from the European Union~\cite{eu_openscience} 
and the Data Sharing Policy from the United States National Institute of Health~\cite{nih_datamanage}. 
Unfortunately, it is out of the scope of this manuscript.

\subsection*{Conclusions}

\ours predicts reactions of given target molecules by predicting their reaction centers, and then completing the 
resulting synthons by attaching small substructures. 
Based on a comparison against twenty baseline methods 
over a benchmark dataset, \ours achieves the state-of-the-art performance under most metrics. 
{The case studies show that} {\ours} also enables diverse predictions. 
However, \ours still has several limitations.
First, the three types of reaction centers in \ours still cannot cover all possible reaction center types
(e.g., the reactions with multiple newly formed bonds). Therefore, a more comprehensive definition of reaction center types
is still needed. 
\ours cannot cover bonds or rings that are attached at the reaction centers but do not appear in the training data either, as 
the substructures that \ours employs to complete synthons are extracted only from training data.
In addition, the atom-mapping between products and reactants that is required by \ours (and required by many existing methods) 
to complete synthons is not always
available or of high quality
(it is available in USPTO-50K). To identify such mappings, it requires to calculate graph isomorphism, which is an 
NP-hard problem. 
Moreover, 
the sum of log-likelihoods of all the involved predictions (i.e., reaction center prediction, attached atom type prediction) 
that \ours uses to prioritize reactions, is not necessarily the same as the likelihood of the reactions, 
which could affect the quality of the prioritized reactions.
We are also investigating a systemic evaluation and \emph{in vitro} validation protocol, 
in addition to using top-$k$ accuracy, as we discussed earlier. 
Multiple-step retrosynthesis could be possible by applying \ours multiple times iteratively, 
each time on a reactant as the target molecule. 
Connected after the deep generative models that have been developed 
to optimize small molecule structures and properties~\cite{Chen2021, jin20a} for lead optimization, 
\ours has a great potential to generate synthetic reactions for these \emph{in silico} generated drug-like molecules, 
and thus substantially speed up the drug development process. 
%

\section*{Methods}
\label{sec:method}

{\ours is developed for the one-step retrosynthesis prediction problem, that is, given the target molecule (i.e., product), 
\ours identifies a set of reactants that can be used to synthesize the molecule through one synthetic reaction.
Following the prior semi-template-based methods~\cite{shi2020g2g,somnath2021learning}, \ours generates reactants from 
products in two steps.
In the first step, \ours identifies the reaction center from the target molecule using the center identification module.
\ours defines the reaction centers as the single bond that is either newly formed or has the bond type changed, or the single atom with changed  
hydrogen count during the reaction.
\ours also incorporates into the reaction center the bonds neighboring the reaction centers 
that have type changes induced by the newly formed bond, 
and the atoms with charge changes within the target molecule (more details in ``Reaction Center Identification" Section). 
Given the reaction center, \ours converts the target molecule into a set of intermediate molecular structures referred to as synthons, 
which are incomplete molecules and will be completed into reactants.
In the second step, \ours completes synthons into reactants by sequentially attaching bonds or rings in the 
synthon completion module. 
The intermediate molecular structures before being completed to reactants 
are referred to as updated synthons.
Figure~\ref{fig:model} presents the overall model architecture of \ours.
All the algorithms are presented in Supplementary Note 5.}

\subsection*{Molecule Representations and Notations}
\label{sec:notations:representation}

Supplementary Table 3 in Supplementary Note 6 presents the key notations used in this manuscript. 
%
%
A synthetic reaction involves a set of reactants $\{\molr\}$ and a product molecule \molp 
that is synthesized from the reactants.
{Please note that we do not consider reagents or catalysts in this study.} 
Each reactant \molr has a corresponding synthon \mols, representing the substructures of \molr that 
appear in \molp.
%
%
We represent the product molecule $\molp$ using a molecular graph $\graphmp$, denoted as $\graphmp=(\atoms, \bonds)$, where $\atoms$ is the set of atoms $\{\atom_i\}$ in 
\molp, and $\bonds$ is the set of corresponding bonds $\{\bond_{ij}\}$, where $\bond_{ij}$ connects atoms $\atom_i$ and $\atom_j$.
We also represent the set of the reactants $\{\molr\}$ or the set of synthons $\{\mols\}$ of \molp using only one molecular graph $\graphmr$ or $\graphms$, 
respectively. Here, \graphmr and \graphms could be disconnected with each connected component representing one reactant or 
one synthon. 

For synthon completion, 
we define a substructure 
\frag as a bond (i.e., $\frag=\bond_{ij}$) or a ring structure 
(i.e., $\frag=\{\bond_{ij}|\atom_i,\atom_j\in \text{a single or polycyclic ring}\}$) that is used to complete synthons into reactants. 
We construct a substructure vocabulary $\vocab=\{\frag\}$ by comparing \graphr's and their corresponding \graphs's in the training data, 
and extracting all the possible substructures from their differences. In total, \ours extracted 83 substructures, 
covering all the reactions in the test data. 
Details about these substructures are available in Supplementary Figure 3 and Supplementary Figure 4 in Supplementary Note 7. 
Note that different from templates used in \tb methods, the substructures \ours used are only bonds and rings, and 
multiple bonds and rings can be attached to complete a synthon. 
For simplicity, when no ambiguity arises, we omit the super/sub-scripts and use {\graph} to represent 
{\graphmm}.

\subsection*{\mbox{Molecule Representation Learning}}
\label{sec:method:representation:learning}
%
\ours learns the atom representations over the molecular graph \graph using the same 
message passing networks (\MPN) as in Chen \etal~\cite{Chen2021} (Supplementary Algorithm 4 in Supplementary Note 5). 

%
\ours first learns atom embeddings to capture the atom types and their local neighborhood structures by passing the messages along the bonds in the molecular graphs. 
Each bond $\bond_{ij}$ is associated with two message vectors $\mess_{ij}$ and $\mess_{ji}$.
The message $\mess_{ij}^{(t)}$ at $t$-th iteration encodes the messages passing from $\atom_i$ to $\atom_j$, and is updated as follows, 
\begin{equation}
\mess_{ij}^{(t)}=W_1^a\relu(W_2^a\vect{x}_i + W_3^a\vect{x}_{ij}+W_4^a\sum_{a_k\in\scriptsize{\neigh}(\atom_i)\backslash\{\atom_j\}}\mess_{ki}^{(t-1)}), 
\end{equation}
where $\vect{x}_i$ is the atom feature vector, including the atom type, valence,  charge, the number of hydrogens, 
whether the atom is included in a ring and whether the ring is aromatic; $\vect{x}_{ij}$ is the bond feature vector, 
including the bond type, whether the bond is conjugated or aromatic, and whether the bond is in a ring; 
$W_i^a$'s ($i$=1,2,3,4) are the learnable parameter matrices; $\mess_{ij}^{(0)}$ is initialized with the zero vector; 
$\neigh(\atom_i)$ is the set with all the neighbors of $\atom_i$ (i.e., atoms connected with $\atom_i$); and 
ReLU is the activation function. 
The message $\mess_{ij}^{(t)}$ captures the structure of $t$-hop neighbors passing through the bond $\bond_{ij}$ to $\atom_j$, 
by iteratively aggregating the neighborhood messages $\mess_{ki}^{(t-1)}$.
With the maximum $t_a$ iterations, \ours derives the atom embedding $\atomEmb_i$ as follows,
\begin{equation}
\label{eqn:atomEmb}
\atomEmb_i=U_1^a\relu(U_2^a\vect{x}_i + U_3^a\sum_{\scriptsize{\atom_k\in\scriptsize{\neigh}(\atom_i)}}\mess_{ki}^{(1 \cdots t_a)}),
\end{equation}
where $\mess_{ki}^{(1\cdots t_a)}$ denotes the concatenation of $\{\mess_{ki}^{(t)}|t\in[1 : t_a]\}$; $U_i^a$'s ($i$=1,2,3) are the learnable parameter matrices.
The embedding of the molecular graph \graph is calculated by summing over all the atom embeddings as follows, 
\begin{equation}
\label{eqn:pool}
\hidden = \sum_{\scriptsize{\atom_i \in \graph}} \atomEmb_i. 
\end{equation}
For \molp and \mols, their embeddings calculated from their moleculear graphs 
as above are denoted as $\hidden_p$ and $\hidden_s$, respectively. 
\subsection*{{Reaction Center Identification}}
\label{sec:method:rci}
%
Given a product $\molp$, \ours defines three types of reaction centers in \molp (Supplementary Algorithm 3 in Supplementary Note 5). 
\begin{enumerate}[noitemsep]
\item  a new bond $\bond_{ij}$, referred to as bond formation center (\bondfmC), that is formed across the reactants during the reaction 
but does not exist in any of the reactants;
\item  an existing bond $\bond_{ij}$ in a reactant, referred to as bond type change center (\bondcgC), whose type changes during the reaction 
due to the gain or loss of hydrogens, while no other changes (e.g., new bond formation) happen; and
\item  an atom in a reactant, referred to as atom reaction center (\atomC), from which a fragment is removed during the reaction, 
without new bond formation or bond type changes. 
\end{enumerate}
The above three types of reaction centers cover 97.7\% of the training set.
The remaining 2.3\% of the reactions in the training data involve multiple new bond formations or bond type changes, 
and will be left for future research. 
Note that with a single atom as the reaction center, the synthon is the product itself.  
We refer to all the transformations needed to change a product to synthons as product-synthon transformations, denoted as \pstrans (Supplementary Algorithm 5 in Supplementary Note 5). 

\subsubsection*{Reaction Centers with New Bond Formation (\bondfmC)}


Following Somnath \etal~\cite{somnath2021learning}, \ours derives the bond representations as follows, 
\begin{equation}
\label{eqn:bondEmb}
\bondEmb_{ij} = U^b_1\relu(U^b_2 \vect{x}_{ij} + U^b_3(\atomEmb_i + \atomEmb_j)+U^b_4 \text{Abs}(\atomEmb_i-\atomEmb_j)),
\end{equation}
where $\text{Abs}(\cdot)$ represents the absolute difference; $U_i^b$'s ($i$=1,2,3,4) are the learnable parameter matrices.
\ours uses the sum and the absolute difference of embeddings of the connected atoms to capture the local neighborhood structure of bond $\bond_{ij}$. 
Meanwhile, the two terms are both permutation-invariant to the order of $\atomEmb_i$ and $\atomEmb_j$, and together can differentiate 
the information in $\atomEmb_i$ and $\atomEmb_j$. 
With the bond representation, \ours calculates a score for each bond $\bond_{ij}$ as follows,
\begin{equation}
\label{eqn:bfcenter}
s^b(\bond_{ij})=\vect{q}^b\relu(Q_1^{b}\bondEmb_{ij} + Q_2^{b}\hidden_{p}),
\end{equation}
where $\hidden_p$ is the representation of the product graph \graphp calculated as in Equation~\ref{eqn:pool}; 
$\vect{q}^b$ is a learnable parameter vector and $Q_1^b$ and $Q_2^b$ are the learnable parameter matrices.
\ours measures how likely bond $\bond_{ij}$ is a \bondfmC using $s^b(\bond_{ij})$ by looking at the bond itself (i.e., $\bondEmb_{ij}$) 
and the structure of the entire product graph (i.e., $\hidden_{p}$).
\ours scores each bond in \molp and selects the most possible \bondfmC candidates $\{\bond_{ij}\}$ with the 
highest scores.
\ours breaks each product at each possible \bondfmC into synthons, and thus can generate multiple
possible reactions. 
%
%


In synthetic reactions, the formation of new bonds could induce the changes of neighbor bonds. 
Therefore, \ours also predicts whether the types of bonds neighboring the \bondfmC 
are changed during the reaction{, referred to as the \bondfmC induced bond type change prediction (\BTCP)}. %
Given the \bondfmC $\bond_{ij}$, the set of the bonds neighboring $\bond_{ij}$ is referred to as 
the \bondfmC neighbor bonds, denoted as \neighCSet, that is: 
\begin{equation}
\neighCSet(\bond_{ij}) = \{\bond_{ik}| \atom_k \in \neigh(\atom_i)\backslash\{\atom_j\}\} \cup \{\bond_{jk}| \atom_k \in \neigh(\atom_j)\backslash\{\atom_i\}\}. 
\end{equation}
Thus, \ours predicts a probability distribution $\mathbf{f}^b\in \mathbb{R}^{1\times4}$ for 
each neighboring bond in \neighCSet, denoted as $\bond_{i/jk} \in \neighCSet$, as follows,
\begin{equation}
\label{eqn:nbchange}
\mathbf{f}^b(\bond_{i/jk})=\text{softmax}(V_1^{b}\bondEmb_{i/jk} + V_2^{b}\bondEmb_{ij}+V_3^{b}\hidden_{p}),
\end{equation}
%
where $V_i^b$'s ($i$=1,2,3) are the learnable parameter matrices.
The first element $\mathbf{f}_1^b$ in $\mathbf{f}^b$ 
represents how likely the $\bond_{i/jk}$ type is changed during the reaction 
(It is determined as type change if $\mathbf{f}^b_1$ is not the maximum in $\mathbf{f}^b$), and the other three 
represent how likely the original $\bond_{i/jk}$ in the reactant is single, double or triple bond, respectively (these three elements
are reset to 0 if $\bond_{i/jk}$ type is predicted unchanged). 
%
%
Here, \ours measures neighbor bond type change by looking at the neighbor bond itself (i.e., $\bondEmb_{i/jk}$), 
the \bondfmC (i.e., $\bondEmb_{ij}$) and the overall product (i.e., $\hidden_{p}$).
\ours updates the synthons \graphs by changing the neighboring bonds of the \bondfmC to their predicted original types.
The predicted changed neighbor bonds are denoted as $\neighCSet^\prime$.
%
%

\subsubsection*{Reaction Centers with Bond Type Change (\bondcgC)} 

If a reaction center is due to a bond type change without new bond formations, 
\ours calculates a score vector $\vect{s}^c\in \mathbb{R}^{1\times3}$ for each bond $\bond_{ij}$ in \molp as follows,
\begin{equation}
\label{eqn:bccenter}
\vect{s}^c(\bond_{ij})=Q_1^c\relu(Q_2^{c}\bondEmb_{ij} + Q_3^{c}\vect{h}_{p}),
\end{equation}
where $Q_i^c$'s ($i$=1,2,3) are the learnable parameter matrices.
%
Each element in $\vect{s}^c(\bond_{ij})$, denoted as $s^c_k(\bond_{ij})$ ($k=1, 2, 3$),  represents, if $\bond_{ij}$ is the \bondcgC, the score of $\bond_{ij}$'s 
original type in \graphr being single, double, and triple bond, respectively.
The element in $\vect{s}^c$ corresponding to  $\bond_{ij}$'s type in \graphp is reset to 0 (i.e., $\bond_{ij}$'s type has to be different 
in \graphr compared to that in \graphp). 
%
%
Thus,  the most possible \bondcgC candidates $\{\bond_{ij}\}$ and their possible original bond types scored by 
$\vect{s}^c(\cdot)$ are selected. 
%
\ours then changes the corresponding bond type to construct the synthons. 

\subsubsection*{Reaction Centers with Single Atoms (\atomC)}

If a reaction center is only at a single atom with a fragment removed, 
\ours predicts a center score for each atom $\atom_i$ in \molp as follows, 
\begin{equation}
\label{eqn:acenter}
s^a(\atom_i)=\vect{q}^{a}\relu(Q_1^{a}\vect{a}_i + Q_2^{a}\hidden_{p}),
\end{equation}
where 
$\vect{q}^a$ is a learnable parameter vector and $Q_1^a$ and $Q_2^a$ are the learnable parameter matrices.
{\ours} selects the atoms $\{\atom_i\}$ in {\molp} with the highest scores as potential \atomC's.
%
In synthon completion, new fragments will be attached at the atom reaction centers.

\subsubsection*{Atom Charge Prediction (\ACP)}


For all the atoms $\atom_i$ involved in the reaction center or \bondfmC changed neighbor bonds $\neighCSet^\prime$, 
\ours also predicts whether the charge of $\atom_i$ 
remains unchanged in reactants. 
%
\ours uses an embedding $\centerEmb$ to represent all the involved bond formations and changes in \pstrans. 
%
%
If the reaction center is predicted as a \bondfmC at $\bond_{ij}$, 
\ours calculates the embedding $\centerEmb$ as follows,  
\begin{equation}
\centerEmb=\sum_{\scriptsize{\bond_{kl}\in \neighCSet^\prime(\bond_{ij})\cup\{\bond_{ij}\}}}W^{c}_1\relu(W^{c}_2\vect{x}^\prime_{kl} + W^{c}_3\bondEmb_{kl}),
\end{equation}
%
where 
$\neighCSet^\prime$ is a subset of \neighCSet with all the bonds that changed types; 
${\vect{x}}^\prime_{kl}$ is a 1$\times$4 one-hot vector,
in which $\vect{x}^\prime_{kl}(0) = 1$ if bond $\bond_{kl}$ is 
 the bond formation center (i.e., $\bond_{kl} = \bond_{ij}$), or $\vect{x}^\prime_{kl}(i) = 1$ ($i$=1, 2, 3) if $\bond_{kl}$ type is changed from 
 single, double or triple bond in reactants, respectively, 
 during the reaction 
(i.e., $\bond_{kl}$ is in $\neighCSet^\prime(\bond_{ij})$);
$W^{c}_i$'s ($i=1,2,3$) are the learnable parameter matrices.
%
%

If the reaction center is predicted as a \bondcgC at $\bond_{ij}$, \centerEmb is calculated as follows, 
\begin{equation}
\centerEmb=W^{c}_1\relu(W^{c}_2\vect{x}^\prime_{ij} + W^{c}_3\bondEmb_{ij}), 
\end{equation} 
where $\vect{x}^\prime_{ij}(0) = 0$ and $\vect{x}^\prime_{ij}(i) = 1$ ($i$=1, 2, 3) if $\bond_{kl}$ type is changed from 
 single, double or triple bond in reactants, respectively,  during the reaction. 
If the reaction center is an \atomC, no \pstrans are needed 
and thus $\centerEmb=\vect{0}$.
%


With the embedding \centerEmb for \pstrans, \ours calculates the probabilities that $\atom_i$
will have charge changes during the reaction
%
as follows,
\begin{equation}
\label{eqn:acchange}
\vect{f}^c(\atom_{i})=\text{softmax}(V_1^{c}\vect{a}_i + V_2^{c}\centerEmb),
\end{equation}
where $V_1^c$ and $V_2^c$ are the learnable parameter matrices; $\vect{f}^c \in \mathbb{R}^{1\times 3}$ is a vector
representing the probabilities of accepting one electron, donating one electron or no electron change during the reaction. 
The option corresponding to the maximum value in $\vect{f}^c$ is selected and will be applied to update synthon charges 
accordingly. 
\ours considers at most one electron change since this is the case for all the reactions in the benchmark data. 
%
%
%

\subsubsection*{Reaction Center Identification Module Training} 

With the scores for three types of reaction centers, 
%
\ours minimizes the following cross entropy loss to learn the above scoring functions (i.e., Equation~\ref{eqn:bfcenter},~\ref{eqn:bccenter} and \ref{eqn:acenter}),
%
\begin{equation}
\mathcal{L}^s = -\sum_{\scriptsize{\bond_{ij}\in \bonds}}\left( y^b_{ij} l^b(\bond_{ij}) + \sum_{k=1}^3 \mathbb{I}_k(y^c_{ij}) l^c_k(\bond_{ij})\right) -\sum_{\scriptsize{\atom_i\in\atoms}}y_i^a l^a(\atom_i),
\end{equation}
%
where $y^*$ ($x=a, b, c$) is the label indicating whether the corresponding candidate 
is the ground-truth reaction center of type $*$ ($y^*=1$) or not ($y^*=0$);
$\mathbb{I}_k(x)$ is an indicator function ($\mathbb{I}_k(x) = 1$ if $x = k$, 0 otherwise), and thus $\mathbb{I}_k(y^c_{ij})$ indicates 
whether the ground-truth bond type of $\bond_{ij}$ is $k$ or not ($k$=1, 2, 3 indicating single, double or triple bond); and 
 $l^*(\cdot)$ ($*=a, b$)/$(l_k^c(\cdot))$  is the probability calculated by normalizing 
 the score $s^*(\cdot)$/$s_k^c$, that is, $l^*(x) = \text{exp}(s^*(x))/\Delta$, 
where $\Delta = \sum_{\scriptsize{\bond_{ij}\in \bonds}}(\text{exp}(s^b(\bond_{ij})) + \sum_{k=1}^3 \text{exp}(s^c_k(\bond_{ij}))) + \sum_{\scriptsize{\atom_i\in\atoms}} \text{exp}(s^a(\atom_i))$
($l^c_k(x) =  \text{exp}(s^c_k(x))/\Delta$). 
%
%
%
Similarly, \ours also learns the predictor $\vect{f}^b(\cdot)$ for neighbor bond changes (Equation~\ref{eqn:nbchange}) 
and $\vect{f}^c(\cdot)$ for atom charge changes (Equation~\ref{eqn:acchange}) by minimizing their respective cross entropy loss 
$\mathcal{L}^b$ and $\mathcal{L}^c$.
%
Therefore, the center identification module learns the predictors by solving the following optimization problem:
\begin{equation}
\min_{\boldsymbol{\Theta}}\mathcal{L}^s + \mathcal{L}^b + \mathcal{L}^c, 
\end{equation}
where $\boldsymbol{\Theta}$ is the set of all the parameters in the prediction functions. 
We used Adam algorithm to solve the optimization problem and do the same for the other training objectives.

\subsection*{Synthon Completion} 
\label{sec:method:synthoncomp}
%

Once the reaction centers are identified and all the product-synthon transformations (\pstrans) are conducted to generate synthons from products, 
\ours completes the synthons into the reactants by sequentially attaching substructures (Supplementary Algorithm 6 in Supplementary Note 5). 
All the actions involved in this process are referred to as synthon-reactant transformations. 
During the completion process, any intermediate molecules $\{\molm\}$ are represented as 
molecular graph $\{\graphm\}$.
%
%
At step $t$, we denote the atom in the intermediate molecular graph $\graphm^{(t)}$ ($\graphm^{(0)}=\graphs$)
that new substructures will be attached to as $\atom^{(t)}$, 
and denote the substructure attached to $\atom^{(t)}$ as $\frag^{(t)}$, resulting in $\graphm^{(t+1)}$.
%

\subsubsection*{Atom Attachment Prediction}

The algorithm for atom attachment prediction is presented in Supplementary Algorithm 8 in Supplementary Note 5. 
%
%
\ours first predicts whether further attachment should be added to $\atom^{(t)}$ or 
should stop at $\atom^{(t)}$, {referred to as the atom attachment continuity prediction (\AACP),}
with the probability calculated as follows,
\begin{equation}
\label{eqn:nfscore}
f^{o}(\atom^{(t)})=\sigma(V_1^{o}\atomEmb^{\scriptsize{(t)}} + V_2^{o}\vect{h}_s + V_3^{o}\vect{h}_p),
\end{equation}
where 
\begin{equation}
\label{eqn:sembedding1}
\vect{h}_s = \sum_{\scriptsize{\atom_i \in \graphs}}\atomEmb_i.
\end{equation}
In Equation~\ref{eqn:nfscore},
 $\atomEmb^{(t)}$ is the embedding of $\atom^{(t)}$ calculated over the graph $\graphm^{(t)}$ (Equation~\ref{eqn:atomEmb}); 
$\vect{h}_s$  is the representation for all the synthons as in Equation~\ref{eqn:sembedding1};   
$V_i^o$'s ($i$=1,2,3) are the learnable parameter matrices;  
$\sigma$ is the sigmoid function.
In Equation~\ref{eqn:sembedding1}, \ours calculates the representations by applying \MPN over the graph $\graphs$ that could be disconnected, 
and the resulted representation is equivalent to applying \MPN over each \graphs's connected component independently and then summing over their representations.
\ours intuitively measures ``how likely'' the atom has a new substructure attached to it by looking at the atom itself (i.e., $\atomEmb^{(t)}$), 
all the synthons (i.e., $\vect{h}_s$), and the product (i.e., $\vect{h}_p$).
Note that in Equation~\ref{eqn:nfscore}, \brics fragment information (i.e., $\atomEmb^\prime$ as in Equation S3 in Supplementary Note 1) is not used
because the fragments for the substructures that will be attached to $\atom^{(t)}$ will not be available until the substructures are determined. 
%


If $\atom^{(t)}$ is predicted to attach with a new substructure, \ours predicts the type of the new substructure, 
{referred to as the atom attachment type prediction (\AATP), }
with the probabilities of all the substructure types in the vocabulary \vocab, calculated as follows, 
\begin{equation}
\label{eqn:fragpred}
\vect{f}^{\scriptsize{\frag}}(\atom^{(t)})=\text{softmax}(V_1^{\scriptsize{\frag}}\atomEmb^{(t)} + V_2^{\scriptsize{\frag}}\vect{h}_s + V_3^{\scriptsize{\frag}}\vect{h}_p),
\end{equation}
where $V_i^{\scriptsize{\frag}}$'s ($i$=1,2,3) are the learnable parameter matrices.
Higher probability for a substructure type $\frag$ indicates that $\frag$ is more likely to be selected as $\frag^{(t)}$.
%
%
The atoms $\atom \in \frag^{(t)}$ in the attached substructure are stored for further attachment, that is, 
they, together with any newly added atoms along the iterative process, 
will become $a^{(T)}$ ($T = t+1, t+2, \cdots$) in a depth-first order in the retrospective reactant graphs.
\ours stops the entire synthon completion process after all the atoms in the reaction centers 
and the newly added atoms are predicted to have 
no more substructures to be attached.

\subsubsection*{Synthon Completion Model Training }

\ours trains the synthon completion module using the teacher forcing strategy, and attaches the ground-truth 
fragments instead of the prediction results to the intermediate molecules during training. 
\ours learns the predictors $f^{o}(\cdot)$ (Equation~\ref{eqn:nfscore})   and 
$\vect{f}^{\scriptsize{\frag}}(\cdot)$ (Equation~\ref{eqn:fragpred})
by minimizing their cross entropy losses 
$\mathcal{L}^{o}$ and $\mathcal{L}^{\scriptsize{\frag}}$ as follows:
\begin{equation}
\min_{\boldsymbol{\Phi}} \mathcal{L}^{o} + \mathcal{L}^{\scriptsize{\frag}}, 
\end{equation}
where $\boldsymbol{\Phi}$ is the set of parameters.

%
\subsection*{Inference}
\label{sec:method:infer}

The algorithm for \ours inference is presented in Supplementary Algorithm 2 in Supplementary Note 5. 

\subsubsection*{Top-$K$ Reaction Center Selection}

During the inference, \ours generates a ranked list of candidate reactant graphs $\{\graphr\}$ 
(note that each reactant 
graph can be disconnected with multiple connected components each representing a reactant).
With a beam size $K$, for each product, 
\ours first selects the top-$K$ most possible reaction centers from each reaction center type (\bondfmC, \bondcgC and \atomC), 
and then selects the top-$K$ most possible reaction centers from all the 3$K$ candidates based on their corresponding scores 
(i.e., $s^b$ as in Equation~\ref{eqn:bfcenter} for \bondfmC, 
$\vect{s}^c$ as in Equation~\ref{eqn:bccenter} for \bondcgC, 
and $s^a$ as in Equation~\ref{eqn:acenter} for \atomC). 
Then \ours converts the product graph \graphp into the top-$K$ synthon graphs $\{\graphsi\}^K_{i=1}$ accordingly. 
{Different reaction centers lead to diverse synthons. }
For these synthon graphs, neighbor bond type change is predicted when necessary; atom charge change is predicted for all the 
atoms involved in reaction centers and their neighboring bonds \neighCSet for \bondfmC's.
%
All the bond type changes and atom charge changes are predicted as those with the highest probabilities as in Equation~\ref{eqn:nbchange} and 
Equation~\ref{eqn:acchange}, respectively. 
%

\subsubsection*{Top-$N$ Reactant Graph Generation}

Once the top-$K$ reaction centers for each product are selected and their synthon graphs are generated, \ours completes the synthon graphs
$\{\graphsi\}^K_{i=1}$ into reactant graphs. During the completion, \ours scores each possible reactant graph and uses their final scores 
to select the top-$N$ reactant graphs, and thus top-$N$ most possible synthetic reactions, for each product. 
Since during synthon completion, the attachment substructure type prediction (Equation~\ref{eqn:fragpred}) gives a distribution of all possible attachment substructures;
by using top possible substructures, each synthon and its intermediate graphs can be extended to multiple different intermediate graphs,
leading to exponentially many reactant graphs {and diversity in the predicted reactions}. 
The intermediate graphs are denoted as $\{\graphm^{(t)}_{ij}\}_{i=1}^K$, where $\graphm^{(t)}_{ij}$ is for the $j$-th possible intermediate graph of the 
$i$-th synthon graph \graphsi at step $t$. 
However, to fully generate all the possible completed reactant graphs, excessive computation is demanded. 
Instead, \ours applies a greedy beam search strategy (Supplementary Algorithm 7 in Supplementary Note 5) 
to only explore the most possible top reactant graph completion paths. 

In the beam search strategy, \ours scores each intermediate graph $\graphm^{(t)}_{ij}$ using a score 
$\score^{(t)}_{ij}$, which is calculated as the sum over all the log-likelihoods of all the 
predictions along the completion path from \graphs up to $\graphm^{(t)}_{ij}$; 
$\score^{(0)}_{ij}$ is initialized as the sum of the log-likelihoods of all the predictions from \graphp to \graphs. 
At each step $t$ ($t \le 30$), each intermediate graph $\graphm_{ij}^{(t)}$ is extended to at most $N$+$1$ intermediate graph candidates. 
These $N$+$1$ candidates
include the one that is predicted to stop at the atom that new substructures could be attached to 
(i.e., as $\atom^{(t)}$ in Equation~\ref{eqn:nfscore}; this intermediate graph could be further completed 
at other atoms)
in this step, 
and at most $N$ candidates with the top-$N$ predicted substructures
attached (Equation~\ref{eqn:fragpred}). 
Among all the candidates generated from all the intermediate graphs at step $t$, 
the top-$N$ scored ones will be further forwarded
into the next completion step $t$+$1$. 
In case some of the top-$N$ graphs are fully completed, the remaining will go through the next steps. 
%
This process will be ended until the number of 
all the completed reactant graphs at different steps reaches or goes above $N$. Then, among all the incomplete graphs at the last step, 
the intermediate graphs with log-likelihood values higher than the $N$-th largest score in all the completed ones will continue to complete as above. 
The entire process will end until 
no more intermediate graphs are qualified to further completion. 
Among all the completed graphs, the top-$N$ graphs are selected as the generated reactants. 

\subsection*{Related Work}
\label{appendix:relate_work}

Deep-learning-based retrosynthesis prediction methods are typically categorized into three classes: template based (\tb), 
template free (\tf) and semi-template based (\semitb).

\subsubsection*{Template-based methods}

Template-based methods formulate the retrosynthesis problem as a selection problem over a set of reaction templates.
These templates can be either hand-crafted by experts~\cite{Szymku2016} or automatically extracted from known reactions in databases~\cite{Coley2017,Segler2017,dai2019,seidl2021modern,chen2019localretro}.
Szymkuc \etal~\cite{Szymku2016} provided a review on using reaction templates coded by human experts for synthetic planning. 
However, these rules may not cover a large set of reactions due to the limitation of human annotation capacity. 
Recent template-based methods extract reaction templates automatically from databases. 
With the reaction templates available, Coley \etal~\cite{Coley2017} (\retrosim) selected the reaction templates 
that the corresponding reactions in the database have the products most similar with the target molecules, 
in order to synthesize the target molecules. 
Dai \etal~\cite{dai2019} learned the joint probabilities of templates matched in the product molecules and all its possible reactants 
using two energy functions, one for reaction template scoring and the other for reactant scoring conditioned on templates. 
Seidl \etal~\cite{seidl2021modern} (\mhnreact) learned to associate the target molecule with the relevant reaction templates using a modern 
Hopfield network.
Chen \etal~\cite{chen2019localretro} (\localretro) scored the suitability of all the reaction templates at all the potential reaction centers 
(atoms and bonds) in the target molecule.
The use of templates provides interpretability towards the reasoning behind the generated reactions.
However, these templates also limit the template-based methods to the reactions only covered by the templates.
%
%
\subsubsection*{Template-free methods}

Template-free methods directly learn to transform the product into the reactants without using the 
reaction templates~\cite{Zheng2019,Chen2019,Mao2021,Irwin2022,Tu2021,Kim2021,sun2021towards,Tetko_2020,Wan2022}.
Most template-free methods utilize the sequence representations of molecules (SMILES) and formulate the transformation 
between the product and its corresponding reactants as a sequence-to-sequence problem.
Many SMILES-based methods use Transformer~\cite{Vaswani2017}, a language model with 
attention mechanisms to model the relationship across tokens.
Transformer follows the encoder-decoder architecture, which encodes the product SMILES 
string into a latent vector and then decodes the vector into the reactant SMILES strings.
For example, Kim \etal~\cite{Kim2021} (\tiedformer) learned the transformation from a product to its reactants using two coupled Transformers with shared parameters, 
one for the forward product prediction (synthesis) and the other for the backward reactant prediction (retrosynthesis).
During the inference, they leveraged both the forward and backward models to find the best reactions. 
Sun \etal~\cite{sun2021towards} (\dual) transformed a product to its reactants using an energy-based framework.
They also leveraged the duality of the forward and backward models by training them together and selected the best reactions 
with the highest energy value from the two models.
{Tetko {\etal}~{\cite{Tetko_2020}} ({\augtrans}) learned to transform a product into its reactants using a Transformer 
	trained on a dataset augmented with various non-canonical SMILES representations of each molecule.
	In {\augtrans}, each target molecule was tested multiple times 
	using different SMILES string representations.
	Zhong \etal~\cite{Zhong2022} (\rsmiles) aligned the product and reactant SMILES strings to minimize their edit distance, and trained 
	a transformer to decode the reactant SMILES strings from the products.
	They also augmented the training dataset and tested each target molecule multiple times as in {\augtrans}.
	In addition to SMILES-based template-free methods, Sacha \etal~\cite{Sacha_2021} (\megan) formulated retrosynthesis as a graph editing process from a product to its reactants.
	These graph edits include the change in the atom properties or the bond types, 
	or the addition of the new atoms or the benzene rings into the synthons.
	These template-free methods are independent of reaction templates, and thus they may have 
	better generalizability to unknown reactions compared to template-based methods.}
However, template-free methods lack interpretability towards the reasoning behind their end-to-end predictions.
SMILES-based template-free methods also suffer from the validity issue that the generated sequences may fail to follow 
the grammar of SMILES strings or violate chemical rules~\cite{Zheng2019}.
%

\subsubsection*{Semi-template-based methods}

Semi-template-based methods~\cite{Wang2021,yan2020retroxpert,shi2020g2g,somnath2021learning,Sacha_2021} 
do not use reaction templates, 
or they do not directly transform a product into its reactants. 
Instead, 
semi-template-based methods 
follow a two-step workflow utilizing atom-mappings: 
(1) {they first} identify the reaction centers and transform the product into synthons (intermediate molecules) using the reaction centers; 
{and then}
(2) {they} complete the synthons into the reactants.
Shi \etal~\cite{shi2020g2g} (\gtog) first predicted reaction centers as bonds that can be used to split the product into the synthons, and 
then utilized {a variational autoencoder}~\cite{kinma2013} to complete synthons 
into reactants by sequentially adding new bonds or new atoms.
Somnath \etal~\cite{somnath2021learning} (\graphretro) predicted the bonds with changed bond types 
or the atoms with changed hydrogen count as the reaction centers, and 
then completed the synthons by selecting the pre-extracted subgraphs that realize the difference 
between synthons and reactants.
Wang \etal~\cite{Wang2021} (\retroprime) formulated the reaction center identification and synthon completion problems 
as two sequence-to-sequence problems (i.e., product to synthon, and synthon to reactant), 
and trained two Transformers for these problems, respectively.
The prediction of reaction centers first in the above methods allows better interpretability towards
the reasoning behind the generation process.
The two-step workflow also empowers these methods to diversify their generated reactants by allowing multiple
different reaction center predictions forwarded into their synthon completion step. 

\ours also identifies the reaction centers and then completes the synthons into the reactants in a sequential way as \gtog does.
{However, \ours is different from \gtog.
	\ours can cover multiple types of reaction centers while \gtog takes only the newly formed bonds as the reaction center, which leads to 
	lower coverage of \gtog on the dataset.
	During synthon completion, \ours attaches substructures (e.g., rings and bonds) 
	instead of single atoms as in \gtog, into synthons to simplify the completion process. 
	%
	In addition and more importantly, \ours uses other synthons {of the same reaction and also the product} 
	to complete a synthon, and thus the synthon completion is more contextualized for the product, 
	while \gtog does not consider other synthons. }

\subsubsection*{Fragment-based molecule generation}

{
	Following the idea of fragment-based drug design~{\cite{Murray2009, hajduk2007decade}}, 
	fragment-based molecule generation methods have been developed. 
	For example, Jin {\etal}~{\cite{jin2018}} 
	first decomposed
	a molecular graph into a junction tree of chemical substructures,
	and then used a variational autoencoder over the junction trees and its chemical substructures to generate and
	assemble new molecules (\jtvae).
	Podda {\etal}~{\cite{podda2020}} encoded and decoded a sequence of fragments via a variational autoencoder, and 
	generated new molecules by connecting fragments generated from the autoencoder. 
	Chen {\etal}~{\cite{Chen2021}} optimized a molecule by removing and attaching substructures in a starting molecule. 
	{\ours} generates reactants from synthons also by attaching new substructures.
	However, the generation strategy in {\ours} is fundamentally different from that in the previous 
	fragment-based molecule generation methods. 
	During synthon completion, {\ours} does not encode the synthons using their substructures as what {\jtvae} and {\modof} do. 
	It does not either encode or decode the substructures that are to be attached to the synthons. 
	Instead, {\ours} attaches the substructures to a specific, identified atom in the molecular graph of the synthons. 
	Therefore, {\ours} can directly attach a substructure to the predicted reaction centers. }

\subsection*{Data Preprocessing and Experimental Settings}
\label{sec:method:data}
%

\begin{table}[!b]
    \caption{\textbf{USPTO-50K data statistics}}
  \label{tbl:stat}
  \centering
  \begin{threeparttable}
      \begin{tabular}{
	@{\hspace{2pt}}l@{\hspace{4pt}}
	@{\hspace{2pt}}l@{\hspace{4pt}}
	@{\hspace{4pt}}r@{\hspace{2pt}}          
	}
        \toprule
        \multicolumn{2}{l}{Dataset}  & Statistics \\
        \midrule
        \multicolumn{2}{l}{\# training reactions} & 40,008 \\
        \multicolumn{2}{l}{\# validation reactions}  & 5,001 \\
        \multicolumn{2}{l}{\# test reactions} & 5,007 \\
        \midrule
        \multirow{4}{*}{\parbox{0.08\textwidth}{training reactions}} 
        & average size of products & 26.0\\
        & average size of larger reactants & 21.9\\
        & average size of smaller reactants  & 9.0\\
        & average number of reactants & 1.7\\
        \midrule
        \multirow{4}{*}{\parbox{0.08\textwidth}{validation reactions}}
        & average size of products & 25.9\\
        & average size of larger reactants & 21.8\\
        & average size of smaller reactants & 9.1\\
        & average number of reactants & 1.7\\
        \midrule
        \multirow{4}{*}{\parbox{0.08\textwidth}{test\\reactions}}
        & average size of products & 25.9\\
        & average size of larger reactants & 21.7\\
        & average size of smaller reactants & 9.2\\
        & average number of reactants & 1.7\\
         \bottomrule
      \end{tabular}
  \end{threeparttable}
  \vspace{-5pt}
\end{table}

We used the benchmark dataset provided by Yan \etal~\cite{yan2020retroxpert}.
This dataset, also referred to as USPTO-50K, contains 50K chemical reactions that are 
randomly sampled from a large dataset collected by Lowe~{\cite{uspto}} 
from US patents published between 1976 and September 2016. 
Each reaction in the large dataset is atom-mapped so that each atom in the product is uniquely 
mapped to an atom in the reactants.
The 50K reactions in USPTO-50K are classified into 10 reaction types by Schneider \etal~\cite{Schneider2016}. 
To avoid the information leakage issue~\cite{yan2020retroxpert} (e.g., reaction center is given in both the training and test data), 
all the product SMILES strings in USPTO-50K are canonicalized. 
We used exactly the same training/validation/test data splits of USPTO-50K as in the previous methods~\cite{Coley2017, yan2020retroxpert}, 
which contain 40K/5K/5K reactions, respectively.
Table~\ref{tbl:stat} presents the data statistics. 
We trained \ours models on the 40K training data, with parameters tuned on the 5K validation data, and tested on the 5K test data. 
For reproducibility purposes, details about model training and parameter tuning are provided in Supplementary Note 8.
%

\subsection*{Baselines}
\label{sec:method:baselines}

We compared \ours with the state-of-the-art baseline methods for the one-step retrosynthesis problem, 
including five template-based (\tb) methods, ten template-free (\tf) methods and five semi-template-based (\semitb) methods.
{
Inspired by the recent success of using fragments in other tasks~\cite{zhang2021}, 
we further extended \ours into \oursb by incorporating the fragments generated from the 
breaking retrosynthetically interesting chemical substructures ({\brics}) fragmentation algorithm~{\cite{Degen2008}}. 
Details of \oursb are available in Supplementary Note 1. 
The experimental setting for \oursb is identical to that of \ours.  
}

\paragraph*{Template-based baseline methods}

The five \tb baseline methods include \retrosim, \neuralsim, \gln, \mhnreact and \localretro. These methods 
first mine reaction templates from training data and apply only these templates to construct reactants from the target molecule.
\begin{itemize}[leftmargin=*,noitemsep]
\item \retrosim~\cite{Coley2017} selects the templates of reactions that produce molecules most similar to the target molecule. 
\item \neuralsim~\cite{Segler2017} predicts suitable templates using product fingerprints through a multi-layer perceptron. 
\item \gln~\cite{dai2019} predicts reactions using two energy functions, one for template scoring and the other for reactant 
scoring conditioned on templates. 
\item \mhnreact~\cite{seidl2021modern} learns the associations between molecules and reaction templates using 
modern Hopfield networks, and selects templates based on the associations. 
\item \localretro~\cite{chen2019localretro} selects templates against each atom and each bond using classifiers. 
\end{itemize}

\paragraph*{{Template-free baseline methods}}

The ten \tf baseline methods all use Transformer
over SMILES string representations of products and/or reactants. 

\begin{itemize}[leftmargin=*,noitemsep]
\item \scrop~\cite{Zheng2019} maps the SMILES strings of products to the SMILES strings of reactants using a Transformer,
and then corrects syntax errors (e.g., mismatch of parentheses in SMILES strings) to ensure valid reactant SMILES strings.
\item \lvtrans~\cite{Chen2019} pre-trains a vanilla Transformer using reactions generated from templates, and then fine-tunes the 
Transformer with a multinomial latent variable representing reaction types. 
\item \get~\cite{Mao2021}  
trains standard Transformer encoders and decoders using the combined atom representations learned from molecular graphs and 
from SMILES strings. 
\item \chemformer~\cite{Irwin2022} translates product SMILES strings into reactant SMILES strings using Transformer, which is 
pre-trained on an independent dataset to recover masked SMILES strings (i.e., with some atoms masked out) 
or to normalize augmented SMILES strings (i.e., multiple, equivalent non-canonical SMILES strings for each SMILES string).  
\item \gtos~\cite{Tu2021} encodes molecular graphs using graph neural networks with attention mechanisms, and 
decodes the reactant SMILES strings from the graph representations using a Transformer decoder.
\item  \tiedformer~\cite{Kim2021} uses two Transformers with shared parameters to learn the transformation from products 
to reactants and vice versa, respectively, and selects the best reactions using the likelihood values from these two Transformers.
\item \gta~\cite{seo2021gta} enhances a Transformer with truncated attention connections regulated by molecular graph structures.  
\item \dual~\cite{sun2021towards} uses an energy-based model with two Transformers to learn the transformation 
from product SMILES strings to reactants' SMILES strings and vice versa, and selects the best reactions using the energy.
\item 
\retroformer~\cite{Wan2022}
{predicts the reaction center region using a reaction center detection module,
and uses the embedding of predicted centers as a condition to transform via Transformer the product into the reactants in SMILES. 
Although {\retroformer} predicts the reaction center, it does not split products into synthons
using the reaction center, and thus does not follow a two-step, semi-template-based framework.}
\item
{{\megan}~\cite{Sacha_2021} transforms the product molecular graphs into the corresponding reactant graphs using
a sequence of graph edits (e.g., change atom charges, add a new bond) that are learned from products and their reactants in the training set. }
\end{itemize}

%

\paragraph*{Semi-template-based methods}

The five \semitb baseline methods all use molecular graph representations. Most of them explicitly predict 
reaction centers first. 

\begin{itemize}[leftmargin=*,noitemsep]
\item 
\retroprime~\cite{Wang2021} trains two Transformers independently to predict the transformation 
from the product to its synthons and from the synthons to the reactants, respectively.
\item
\retroxpert~\cite{yan2020retroxpert} predicts reaction centers on molecular graphs via a graph attention network, 
and transforms resulting synthons to reactants using a Transformer. 
\item
\gtog~\cite{shi2020g2g} predicts reaction centers on molecular graphs via a graph neural network, and completes synthons
into reactants through sequential additions of new atoms or bonds using the latent variables sampled from the latent space of 
a variational graph autoencoder. 
\item
\graphretro~\cite{somnath2021learning} predicts reaction centers via a message passing neural network over molecular graphs,
and completes synthons by selecting
the subgraphs in a vocabulary that realize the difference between the synthons and reactants.
\end{itemize}

\section*{Data Availability}
\label{sec:data_availability}

The data used in this manuscript are available publicly~\cite{g2retro_code} at the link \url{https://doi.org/10.5281/zenodo.7839013} and the link \url{https://github.com/ninglab/G2Retro}.

\section*{Code Availability}
\label{sec:code_availability}

{The code for \ours, \oursb and \oursens is available publicly~\cite{g2retro_code} at the link \url{https://doi.org/10.5281/zenodo.7839013} and the link \url{https://github.com/ninglab/G2Retro}.
{A web portal for \ours is available at the link \url{http://go.osu.edu/G2Retro}.}

\section*{Acknowledgements}

This project was made possible, in part, by support from the National Science
Foundation grant nos.
IIS-2133650 (X.N.), and The Ohio State University President's Research Excellence
program (X.N., H.S.). Any opinions, findings and conclusions or recommendations
expressed in this paper are those of the authors and do not necessarily reflect the views of the funding agency. 
We thank Dr. Michael A. Walters for his constructive comments.

\section*{Author Contributions}

X.N. conceived the research. X.N. and H.S. obtained funding for the research. 
Z.C. and X.N. designed the research. Z.C. and
X.N. conducted the research, including data curation, formal analysis, methodology
design and implementation, result analysis and visualization. Z.C. and X.N. drafted the original
manuscript. 
O.R.A. and J.R.F. provided comments on case studies. 
H.S. provided comments on the original manuscript. Z.C. and X.N.
 conducted the manuscript editing and revision. All authors reviewed the final
manuscript.

\section*{Competing Interests}

The authors declare no competing interests.

\clearpage

\setcounter{section}{0}
\renewcommand{\thesection}{Supplementary Note \arabic{section}: }

\setcounter{table}{0}
\renewcommand{\thetable}{\arabic{table}}

\setcounter{figure}{0}
\renewcommand{\thefigure}{\arabic{figure}}

\setcounter{algorithm}{0}
\renewcommand{\thealgorithm}{\arabic{algorithm}}

\setcounter{equation}{0}
\renewcommand{\theequation}{S\arabic{equation}}

\renewcommand{\figurename}{Supplementary Figure}

\renewcommand{\tablename}{Supplementary Table}
\renewcommand{\refname}{Supplementary References}

\makeatletter
\newcommand{\newalgname}[1]{%
  \renewcommand{\ALG@name}{#1}%
}
\newalgname{Supplementary Algorithm}
\renewcommand{\listalgorithmname}{Liste des \ALG@name s}
\makeatother

\makeatletter
\renewcommand\@biblabel[1]{#1.}
\makeatother

\makeatletter
\renewenvironment{table*}%
{\renewcommand\familydefault\sfdefault
	\@float{table}}
{\end@float}
\makeatother

\begin{center}
\begin{minipage}{0.95\linewidth}
	\centering
	\LARGE 
	\ours as a Two-Step Graph Generative Models for Retrosynthesis Prediction (Supplementary Information)
\end{minipage}
\end{center}
\vspace{10pt}

\vspace{-2\baselineskip}

\section{{\ours with fragments: \oursb}}
\label{appendix:G2RetroB}

{
Inspired by the recent success of using fragments in other tasks~\cite{zhang2021}, 
we extended \ours by incorporating the fragments generated from the 
breaking retrosynthetically interesting chemical substructures ({\brics}) fragmentation algorithm~{\cite{Degen2008}}, 
and denote the new method as \oursb. 
{\brics} breaks synthetically accessible bonds in a product {\molp}, following a set of fragmentation rules.
Thus, the fragments generated from \brics encode prior knowledge related to synthesis. 
\oursb integrates such knowledge by learning from the molecular graph constructed from the fragments. 
Specifically, for each \molp, 
{\oursb} constructs a {\brics} graph $\treep=(\vertices, \edges)$, 
where each node $\node_u \in \vertices$ represents
a {\brics} fragment with all the atoms and bonds belonging to it, 
and each edge $\edge_{uv} \in \edges$ corresponds to a bond $\bond_{ij}$ 
that connects two {\brics} fragments $\node_u$ and $\node_v$. That is, the two atoms connected by $\bond_{ij}$ belong 
to $\node_u$ and $\node_v$, respectively (i.e., $\atom_i \in \node_u$ and $\atom_j \in \node_v$; 
In our dataset, two {\brics} fragments are connected through only one bond). 
Thus, {\edges} includes synthetically accessible bonds, which tend to be the reaction centers, 
and thus, {\treep} incorporates fragment-level structures of {\molp}.  
For simplicity, when no ambiguity arises, we omit the super/sub-scripts and use {\tree} to represent 
{\treep}. }

{\oursb} generates {\brics} fragment embeddings by passing the messages along the connections over {\brics} fragments in the {\brics} graphs, 
in a similar way as for atom embeddings over molecular graphs. 
Specifically, each edge $\edge_{uv}$ in \tree 
is associated with two message vectors $\vect{e}_{uv}$ and $\vect{e}_{vu}$.
The message $\vect{e}_{uv}^{(t)}$ at $t$-th iteration is updated as follows, 
\begin{equation}
\edgeEmb_{uv}^{(t)}=W_1^e\relu(W_2^e\vect{s}_u + W_3^e\vect{s}_{uv}+W_4^e\sum_{\scriptsize{\node_w\in\scriptsize{\neigh}(\node_u)\backslash \{\node_v\}}}\edgeEmb_{wu}^{(t-1)}),
\end{equation}
where $\vect{s}_u=\sum\nolimits_{\scriptsize{\atom_i\in \node_u}}\vect{a}_i$ aggregates the embeddings of all the atoms within the fragment $\node_u$;
$\vect{s}_{uv}= \atomEmb_i$ is the embedding of atom $\atom_i$ in $\node_u$ that is included in the edge $\edge_{uv}$;
$W_i^e$'s ($i$=1,2,3,4) are the learnable parameter matrices; 
$\vect{e}_{uv}^{(0)}$ is initialized with the zero vector.
The message $\vect{e}_{uv}^{(t)}$ encodes the information passing through the edge $\edge_{uv}$ to $\node_v$, and thus is used to further derive the embedding of $\node_v$ as follows,
\begin{equation}
\nodeEmb_v=U_1^e\relu(U_2^e\vect{s}_v + U_3^e\sum_{\scriptsize{\node_w\in\scriptsize{\neigh}(\node_v)}}\edgeEmb_{wv}^{(1 \cdots t_e)}),
\end{equation}
where $\edgeEmb_{wv}^{(1\cdots t_e)}$ denotes the concatenation of $\{\edgeEmb_{wv}^{(t)}|t\in[1 : t_e]\}$; $U_i^e$'s ($i$=1,2,3) are the learnable parameter matrices.
With \tree, \oursb enriches the representation of atom $\atom_i$ with the embedding $\nodeEmb_v$ of the fragment that $\atom_i$ belongs to. 
Note that in \brics algorithm, each atom only belongs to one fragment. 
The enriched atom representation is calculated as follows, 
\begin{equation}
\label{eqn:enrich}
{\atomEmb}^\prime_i=V(\atomEmb_i\oplus\nodeEmb_v), 
\end{equation}
where $V$ is a learnable hyperparameter matrix; $\oplus$ represents the concatenation operation.

{
The reaction center identification in \oursb is done in the same way as in \ours (Section ``Reaction Center Identification" in 
the main manuscript), with all the enriched atom representations calculated as above, 
and bond embeddings (e.g., Equation~\ref{eqn:bondEmb} in the main manuscript) calculated using 
the enriched atom representations.
Note that synthon completion in \oursb does not use the {\brics} graph and thus is identical to {\ours}. 
} 
%

\section{{{\ours} ensemble: {\oursens}}}
\label{appendix:ensemble}

{
To explore a large reaction space, we developed an ensemble approach for \ours, denoted as \oursens. 
\oursens ensembles 20 \ours models that are combined from the top-4 reaction center identification modules 
and the top-5 synthon completion modules, each selected based on the corresponding validation data
(hyper-parameter space follows that in Supplementary Table~\ref{tbl:parameter} except for atom embedding dimensions in \{32, 64\}).  
For each target product, all the top-10 predicted reactions from the 20 \ours models are combined based on their 
average ranking (different \ours may predict the same reaction), and the final top-10 predicted reactions are considered 
as the results of \oursens. 
}

\begin{table*}[t]
	\captionof{table}{\textbf{{Performance comparison between \oursens and \rsmiles on different reaction types}}}
	\label{tbl:class_aug}
  \vspace{-8pt}    
\begin{center}
\begin{small}
\begin{threeparttable}
	\begin{tabular}{
			@{\hspace{2pt}}l@{\hspace{5pt}}
			@{\hspace{2pt}}r@{\hspace{5pt}}
			@{\hspace{5pt}}r@{\hspace{5pt}}
			@{\hspace{5pt}}c@{\hspace{5pt}}
		    	@{\hspace{5pt}}c@{\hspace{5pt}}
			@{\hspace{5pt}}c@{\hspace{5pt}}
			@{\hspace{5pt}}c@{\hspace{5pt}}
			@{\hspace{2pt}}c@{\hspace{2pt}}
			%
			@{\hspace{5pt}}c@{\hspace{5pt}}
			@{\hspace{5pt}}c@{\hspace{5pt}}
			@{\hspace{5pt}}c@{\hspace{5pt}}
			@{\hspace{5pt}}c@{\hspace{5pt}}
			%
		}
			\toprule
			 \multirow{2}{*}{Type Name} & \multirow{2}{*}{\parbox{0.05\textwidth}{Percentage (\%)}} &  & \multicolumn{4}{c}{\oursens} & & \multicolumn{4}{c}{\rsmiles}\\
			\cmidrule(lr){4-7} \cmidrule(lr){9-12}
			& & & 1 & 3 & 5 & 10 & &  1 & 3 & 5 & 10 \\
			\midrule 
			heteroatom alkylation and arylation    &  30.3 &&  56.5 &  80.7 &  88.5 &  94.4 & & 56.5 &  81.3 &  88.1 &  93.5 \\
	       	acylation and related processes        &  23.8 &&  69.7 &  91.1 &  95.4 &  98.2 & & 68.7 &  89.8 &  93.9 &  96.4 \\
	       	deprotections                          &  16.5 &&  54.2 &  80.1 &  87.1 &  92.4 & & 52.7 &  76.6 &  81.4 &  86.7 \\
			C-C bond formation                     &  11.3 &&  41.4 &  64.2 &  71.4 &  80.6 & & 39.7 &  63.5 &  74.3 &  81.7 \\
			reductions                             &   9.2 &&  61.0 &  78.4 &  84.6 &  90.7 & & 59.3 &  80.1 &  87.7 &  92.2 \\
			functional group interconversion      &   3.7  &&  35.3 &  57.1 &  67.4 &  73.4 & & 42.4 &  57.6 &  66.3 &  79.3 \\
			heterocycle formation                  &  1.8  &&   0.0 &   0.0 &   0.0 &   0.0 & & 48.4 &  70.3 &  78.0 &  83.5 \\
			oxidations                             &   1.6 &&  68.3 &  86.6 &  90.2 &  93.9 & & 54.9 &  82.9 &  92.7 &  97.6 \\
			protections                            &  1.4  &&  51.5 &  77.9 &  85.3 &  89.7 & & 58.8 &  82.4 &  88.2 &  91.2 \\
			functional group addition             &   0.5  &&  78.3 &  87.0 &  87.0 &  95.7 & & 78.3 &  87.0 &  91.3 &  95.7 \\
			\bottomrule
	\end{tabular}%
	\begin{tablenotes}[normal,flushleft]
	\begin{footnotesize}
	\item {\!Columns with 1, 3, 5 and 10 present top-1, top-3, top-5 and top-10 accuracies, respectively. }
	{Column ``Percentage(\%)'' represents the percentage of reactions in the test set belonging to the specific reaction type.}
		\par 
		\end{footnotesize}
	\end{tablenotes}
	%
\end{threeparttable}
\end{small}
  \vspace{-10pt}    
\end{center}
\end{table*}

{
Supplementary Table~\ref{tbl:class_aug} presents the performance comparison between \oursens and \rsmiles on different reaction types.
Among the 10 reaction types in the benchmark data, \oursens outperforms \rsmiles at top-1 accuracy on 5 reaction types, 
and archives the same performance on 2 reaction types. On average, \oursens outperforms \rsmiles on the most popular
reaction types on higher-ranked predictions (i.e., corresponding to smaller $k$ in top-$k$ accuracy).
For example,  \oursens substantially outperforms \rsmiles on deprotection reactions (54.2\% vs 52.7\% on top-1 accuracy).
}

\section{{Additional case study}}
\label{appendix:case_study}

%
For Mavacamten as in Supplementary Figure~\textbf{1a}, 
which was approved by FDA in 2022 to treat hypertrophic cardiomyopathy~\cite{Keam_2022_mavacamten}, 
the patent literature~\cite{mavacamten} 
reports the utilization of a nucleophilic aromatic substitution for the formation of the C8-N9 bond 
(ground truth in Supplementary Figure~\textbf{1b}).
\ours correctly predicts this coupling as the top-1 reaction (Supplementary Figure~\textbf{1c}), 
and also identifies its additional permutations by replacing the aryl chloride with the aryl bromide and the aryl fluoride, respectively (Supplementary Figure~\textbf{1d} and \textbf{1g}). 
Aryl fluorides in Supplementary Figure~\textbf{1g} 
are not as typical as aryl chlorides and bromides, and \ours ranks the substitution reaction involving the arly fluoride low.
In addition to the amine coupling strategy with aryl halides, \ours also identifies the reaction of the amine with trifluoro methyl sulfate to make the same bond (Supplementary Figure~\textbf{1h}), 
which would be expected to work as with aryl halides in Supplementary Figure~\textbf{1c} and \textbf{1d}. 
However, the alcohol in Supplementary Figure~\textbf{1i} 
is not a good enough leaving group to make the bond (i.e., C8-N9).
Interestingly, \ours also identifies other amine linkages (e.g., in Supplementary Figure~\textbf{1e} 
between C2 and N4; in Supplementary Figure~\textbf{1j} 
between N9 and C10) as potential reaction centers.
However, the proposed synthesis in Supplementary Figure~\textbf{1e} and \textbf{1f} 
would most likely lead to the formation of undesired products as these reactant pairs would likely result in the alkylation of both N4 and N9.
Therefore, the use of the aryl halides in Supplementary Figure~\textbf{1c} and \textbf{1d} 
would be the more efficient way of obtaining the desired product.

{Oteseconazole as in Supplementary Figure~\textbf{2a} 
is a drug approved for recurrent vulvovaginal candidiasis~{\cite{Hoy2022}}.
In the patent literature~{\cite{oteseconazole}}, this drug is constructed by the C-C bond forming reaction between C6 and C7 
and is assembled with Suzuki coupling~{\cite{Miyaura1995}} between an aryl bromide group and a boronic ester (Supplementary Figure~\textbf{2b}). 
{\ours} correctly predicts this coupling as the top-1 with the boronic acid (Supplementary Figure~\textbf{2c}), 
the top-3 which is the same as the patented reaction (Supplementary Figure~\textbf{2e}), 
and the top-9 reaction with a relatively uncommon boronic ester (Supplementary Figure~\textbf{2k}). 
Boronic acids in Supplementary Figure~\textbf{2c} 
would typically be considered by synthetic chemists as interchangeable with boronic esters, and thus should be considered a feasible reaction;
while Boronic ester in Supplementary Figure~\textbf{2k} 
should react in the same way with the patented reaction, {and thus could deliver the desired compound.}
Interestingly, {\ours} also predicts the Ullmann-type coupling~{\cite{FANTA1974}} with different aryl halides to construct the C6-C7 bond in Supplementary Figure~\textbf{2d}, \textbf{2f} and \textbf{2g}, 
all of which would be expected as feasible reactions.
{Although the reaction center is correctly identified in Supplementary Figure~\textbf{2i}, 
the proposed coupling of two boronic acids would not be effective.}
%
{\ours} also identifies another C-N coupling of various aryl halides with imidazoles (C15-N16 - Supplementary Figure~\textbf{2h} and \textbf{2j}), 
which hypothetically would also work as expected.
}

\begin{figure*}[!h]
	\centering
	\begin{subfigure}[t]{0.24\textwidth}
		\centering
		\includegraphics[width=0.78\linewidth]{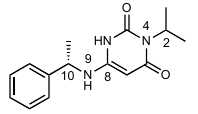}
		\caption*{\textbf{1a}, Mavacamten}	
		\label{fig:mavacamten:product}
	\end{subfigure}
	\begin{subfigure}[t]{0.24\textwidth}
		\centering
		\includegraphics[width=0.95\linewidth]{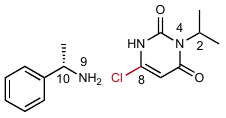}
		\caption*{\textbf{1b}, ground-truth reactants}
		\label{fig:mavacamten:gt}
	\end{subfigure}
	\begin{subfigure}[t]{0.24\textwidth}
		\centering
		\includegraphics[width=0.95\linewidth]{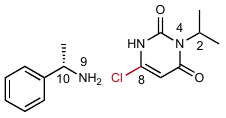}
		\caption*{\textbf{1c}, top-1 predicted reactants}
		\label{fig:mavacamten:top1}
	\end{subfigure}
	\begin{subfigure}[t]{0.24\textwidth}
		\centering
		\includegraphics[width=0.95\linewidth]{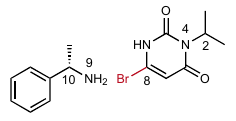}
		\caption*{\textbf{1d}, top-2 predicted reactants}
		\label{fig:mavacamten:top2}
	\end{subfigure}
	\\
	\begin{subfigure}[t]{0.24\textwidth}
		\centering
		\includegraphics[width=0.98\linewidth]{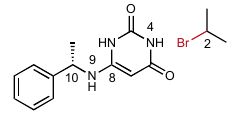}
		\caption*{\textbf{1e}, top-3 predicted reactants}
		\label{fig:mavacamten:top3}
	\end{subfigure}
	\begin{subfigure}[t]{0.24\textwidth}
		\centering
		\includegraphics[width=0.98\linewidth]{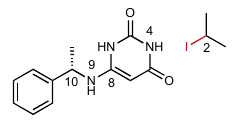}
		\caption*{\textbf{1f}, top-4 predicted reactants}
		\label{fig:mavacamten:top4}
	\end{subfigure}
	\begin{subfigure}[t]{0.24\textwidth}
		\centering
		\includegraphics[width=0.95\linewidth]{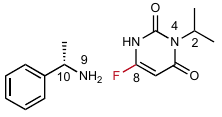}
		\caption*{\textbf{1g}, top-5 predicted reactants}
		\label{fig:mavacamten:top5}
	\end{subfigure}
	\begin{subfigure}[t]{0.24\textwidth}
		\centering
		\includegraphics[width=\linewidth]{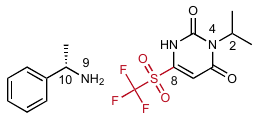}
		\caption*{\textbf{1h}, top-6 predicted reactants}
		\label{fig:mavacamten:top6}
	\end{subfigure}
	\\
	\vspace{-2pt}
	\begin{subfigure}[t]{0.24\textwidth}
		\centering
		\includegraphics[width=0.95\linewidth]{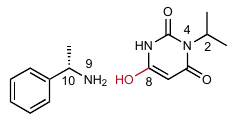}
		\caption*{\textbf{1i}, top-7 predicted reactants}
		\label{fig:mavacamten:top7}
	\end{subfigure}
	\begin{subfigure}[t]{0.24\textwidth}
		\centering
		\includegraphics[width=0.95\linewidth]{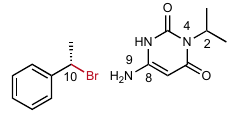}
		\caption*{\textbf{1j}, top-8 predicted reactants}
		\label{fig:mavacamten:top8}
	\end{subfigure}
	\begin{subfigure}[t]{0.24\textwidth}
		\centering
		\includegraphics[width=0.95\linewidth]{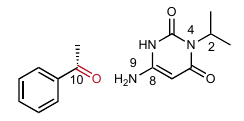}
		\caption*{\textbf{1k}, top-9 predicted reactants}
		\label{fig:mavacamten:top9}
	\end{subfigure}
	\begin{subfigure}[t]{0.24\textwidth}
		\centering
		\includegraphics[width=0.78\linewidth]{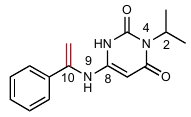}
		\caption*{\textbf{1l}, top-10 predicted reactants}
		\label{fig:mavacamten:top10}
	\end{subfigure}
	\vspace{-5pt}
	\caption{\textbf{Predicted reactions by \ours for ``mavacamten''}. 
		Numbers next to each atom are the indices of the atoms. Atoms with same indices 
		in different subfigures are corresponding to each other. 
		Atoms and bonds colored in red are leaving groups for synthon completion.
		\textbf{1a}, product/target molecule; \textbf{1b}, the ground-truth reactants in USPTO-50K; 
		\textbf{1c-1l}, top predicted reactants.}
	\label{fig:mavacamten}
\end{figure*}

\begin{figure*}[!h]
	\centering
	\begin{subfigure}[t]{0.32\textwidth}
		\centering
		\includegraphics[width=0.9\linewidth]{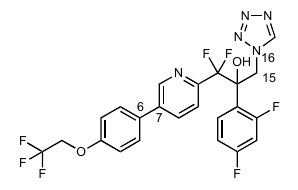}
		\caption*{\textbf{2a}, Oteseconazole}
		\label{fig:oteseconazole:product}
	\end{subfigure}
	\begin{subfigure}[t]{0.32\textwidth}
		\centering
		\includegraphics[width=0.95\linewidth]{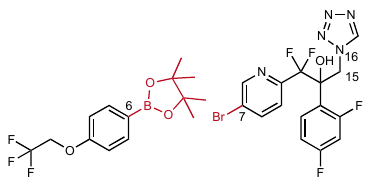}
		\caption*{\textbf{2b}, ground-truth reactants}
		\label{fig:oteseconazole:gt}
	\end{subfigure}
	\begin{subfigure}[t]{0.32\textwidth}
		\centering
		\includegraphics[width=0.95\linewidth]{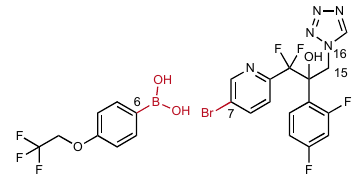}
		\caption*{\textbf{2c}, top-1 predicted reactants}
		\label{fig:oteseconazole:top1}
	\end{subfigure}
	\\
	\begin{subfigure}[t]{0.32\textwidth}
		\centering
		\includegraphics[width=0.95\linewidth]{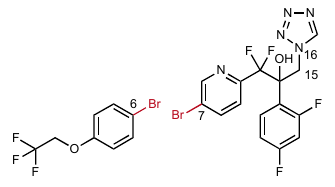}
		\caption*{\textbf{2d}, top-2 predicted reactants}
		\label{fig:oteseconazole:top2}
	\end{subfigure}
	%
	\begin{subfigure}[t]{0.32\textwidth}
		\centering
		\includegraphics[width=0.95\linewidth]{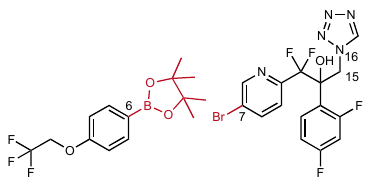}
		\caption*{\textbf{2e}, top-3 predicted reactants}
		\label{fig:oteseconazole:top3}
	\end{subfigure}
	\begin{subfigure}[t]{0.32\textwidth}
		\centering
		\includegraphics[width=0.95\linewidth]{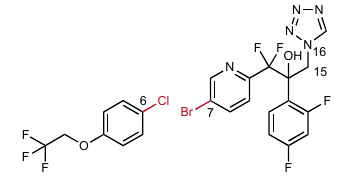}
		\caption*{\textbf{2f}, top-4 predicted reactants}
		\label{fig:oteseconazole:top4}
	\end{subfigure}
	\\
	\begin{subfigure}[t]{0.32\textwidth}
		\centering
		\includegraphics[width=0.95\linewidth]{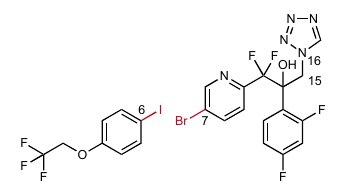}
		\caption*{\textbf{2g}, top-5 predicted reactants}
		\label{fig:oteseconazole:top5}
	\end{subfigure}
	\begin{subfigure}[t]{0.32\textwidth}
		\centering
		\includegraphics[width=0.95\linewidth]{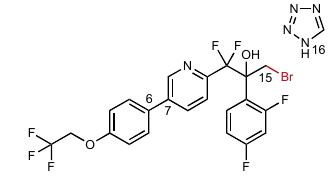}
		\caption*{\textbf{2h}, top-6 predicted reactants}
		\label{fig:oteseconazole:top6}
	\end{subfigure}
	\begin{subfigure}[t]{0.32\textwidth}
		\centering
		\includegraphics[width=0.95\linewidth]{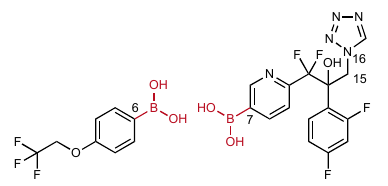}
		\caption*{\textbf{2i}, top-7 predicted reactants}
		\label{fig:oteseconazole:top7}
	\end{subfigure}
	\\
	\begin{subfigure}[t]{0.32\textwidth}
		\centering
		\includegraphics[width=0.95\linewidth]{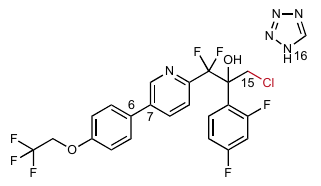}
		\caption*{\textbf{2j}, top-8 predicted reactants}
		\label{fig:oteseconazole:top8}
	\end{subfigure}
	\begin{subfigure}[t]{0.32\textwidth}
		\centering
		\includegraphics[width=0.95\linewidth]{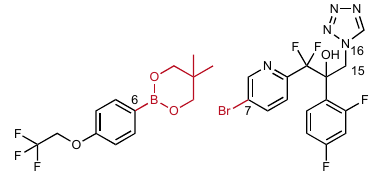}
		\caption*{\textbf{2k}, top-9 predicted reactants}
		\label{fig:oteseconazole:top9}
	\end{subfigure}
	\begin{subfigure}[t]{0.32\textwidth}
		\centering
		\includegraphics[width=0.95\linewidth]{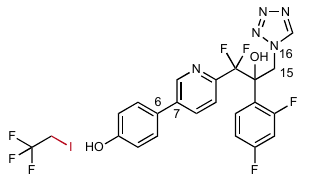}
		\caption*{\textbf{2l}, top-10 predicted reactants}
		\label{fig:oteseconazole:top10}
	\end{subfigure}
    \vspace{-10pt}
	\caption{\textbf{Predicted reactions by \ours for Oteseconazole}.
		Numbers next to each atom are the indices of the atoms. Atoms with same indices 
		in different subfigures are corresponding to each other. 
		Atoms and bonds colored in red are leaving groups for synthon completion.
		\textbf{2a}, product/target molecule; \textbf{2b}, the ground-truth reactants in USPTO-50K; 
		\textbf{2c-2l}, top predicted reactants.}
	\label{fig:oteseconazole}
\end{figure*}

\section{{Clustering algorithms for diversity analysis}}
\label{appendix:cluster}

{Supplementary Algorithm~{\ref{alg:cluster}} describes the algorithm to cluster products for diversity analysis. 
Given $K$ products $\{\mol_{p}^k\}_{k=1, \cdots, K}$ and their top-10 predicted reactions $\{\{R_{i}^k\}_{i=1,\cdots, 10}\}_{k=1, \cdots, K}$, we clustered products according to their reaction 
similarity distributions.
}

\begin{algorithm}[!h]
	\caption{{Clustering products according to reaction similarity distributions}}
	\label{alg:cluster}
	\begin{algorithmic}[1]
		\Require $\{\mol_{p}^k, \{R_{i}^k\}_{i=1, \cdots, 10}\}_{k=1, \cdots, K}$, NClusters
		\vspace{0.3em}
		\For{each $\mol_{p}^k$ and $\{R_{i}^k\}_{i=1, \cdots, 10}$}
		\ShortLineComment{calculate pair-wise similarities among top-10 predictions (Equation~\ref{eqn:sim})}
		\For{each pair of reactions $(R_i^k, R_j^k)$ with $i\neq j$}
		\State $s_{ij}^k = \text{sim}(R_i^k, R_j^k)$
		\EndFor
		\ShortLineComment{generate the reaction similarity distribution of product $\mol_{p}^k$}
		\State $\vect{h}^k = \text{histogram}(\{s_{ij}^k\}_{\forall i, j})$
		\EndFor
		\LineComment{cluster products using K-Means according to their reaction similarity distributions}
		\State $\{C_i\}_{i=1, \cdots, \text{NClusters}}=\text{K-Means}(\{\vect{h}^k\}_{k=1, \cdots, K}, \text{NClusters})$
		\State \Return  $\{C_i\}_{i=1, \cdots, \text{NClusters}}$
	\end{algorithmic}
\end{algorithm}

\section{Algorithms of \ours}
\label{appendix:pseudo_code}

Supplementary Algorithm~\ref{alg:infer} describes the reactant generation process of \ours.
Given a product, the maximum number of synthons $K$, the beam size $N$, and the maximum number of steps allowed maxSteps, 
\ours generate a ranked list of $N$ reactants that can be used to synthesize the product.
Supplementary Algorithm~\ref{alg:RCI} describes how \ours converts the product graph into top-$K$ synthon graphs.
Given a product graph $\graphp$, its corresponding \brics graph $\tree$ and $K$, \ours
predicts the top-$K$ synthon graphs and calculates their log-likelihood scores $\{\score_k\}$, 
using the learned molecule representations from the encoder described in Supplementary Algorithm~\ref{alg:encoder}.
Specifically, \ours first selects the top-$K$ most possible reaction centers and calculates their log-likelihood scores
$\{\score_k, \rcenter_k\}^K_{k=1}$.
Then given the product graph, the top-$K$ reaction centers and their scores and the product molecule
representation $\hidden_{p}$, \ours transforms the product graph into top-$K$ synthon graphs as in Supplementary Algorithm~\ref{alg:p2st}.
Supplementary Algorithm~\ref{alg:syncomp} describes how \ours completes top-$K$ synthon graphs into top-$N$ reactant graphs.
Given the product graph $\graphp$, the top-$K$ synthon graphs and their scores $\{\score_k, \graph_{s,k}\}^K_{k=1}$, 
the beam size $N$, and the maximum number of completion steps maxSteps,
\ours uses a beam search strategy to complete the synthon graphs into the reactant graphs in a sequential way.
Supplementary Algorithm~\ref{alg:beamsearch} describes the beam search strategy.
Given the queue of intermediate molecules $Q$, the queue of completed reactants $R$, 
the representations of top-$K$ synthons $\{\hidden_{s,k}\}^K_{k=1}$, the product representation $\hidden_p$,
and the beam size $N$, \ours extends each intermediate molecule in the queue by attaching different substructures 
at the attachment point, and saves the completed molecules into $R$ and the incomplete ones for the next completion step. 
Supplementary Algorithm~\ref{alg:aap} describes how to attach new substructures to an intermediate molecule using the Atom Attachment Continuity Prediction (\AACP) and the Atom Attachment Type Prediction (\AATP).

\begin{algorithm}[!h]
	\caption{\ours}
	\label{alg:infer}
	\begin{algorithmic}[1]
		\Require $\molp = (\graphp, \graph^B_p)$, $K$, $N$, maxSteps
		\LineComment{predict top-$K$ synthons with Supplementary Algorithm~\ref{alg:RCI}}
		\State $\{\score_k,\graph_{s,k}\}_{k=1}^K = \ours\text{-\RCI}(\graphp, \graph^B_p, K)$
		\vspace{0.3em}
		\LineComment{predict top-$N$ reactants with Supplementary Algorithm~\ref{alg:syncomp}}
		\State $\{\graph_{r,i}\}_{i=1}^N = \ours\text{-\syncomp}(\graphp, \{\score_k,\graph_{s,k}\}_{k=1}^K, N, \text{maxSteps})$
		\vspace{0.3em}
		\State \Return  $\{\graph_{r,i}\}_{i=1}^N$
	\end{algorithmic}
\end{algorithm}

\begin{algorithm}[!h]
	\caption{\ours-\RCI for Reaction Center Identification}
	\label{alg:RCI}
	\begin{algorithmic}[1]
		\Require \graphp, \tree, $K$
		\vspace{0.3em}
		\LineComment{learn molecule representations with Supplementary Algorithm~\ref{alg:encoder}}
		\State $\{\atomEmb_i\},\{\bondEmb_{ij}\}, \hidden_p=\ours\text{-encoder}(\graphp, \tree)$
		\vspace{0.3em}
		\LineComment{select top-$K$ $\mathsf{BF\text{-}centers}$ (Equation~\ref{eqn:bfcenter})}
		\State $\{s^b(\bond_{ij})\}^K=\text{top}(K, \text{findCenter}(\bondfmC, \{\bondEmb_{ij}\}, \hidden_{p})\text{)}$
		\vspace{0.3em}
		\LineComment{select top-$K$ $\mathsf{BC\text{-}centers}$ (Equation~\ref{eqn:bccenter})}
		\State $\{s^c_k(\bond_{ij})\}^K=\text{top}(K, \text{findCenter}(\bondcgC, \{\bondEmb_{ij}\}, \hidden_{p})\text{)}$
		\vspace{0.3em}
		\LineComment{select top-$K$ $\mathsf{A\text{-}centers}$ (Equation~\ref{eqn:acenter})}
		\State $\{s^a(\atom_{i})\}^K=\text{top}(K, \text{findCenter}(\atomC, \{\atomEmb_{i}\}, \hidden_{p})\text{)}$
		\vspace{0.3em}
		\LineComment{select top-$K$ centers $\{\rcenter_k\}$ and calculate their log-likelihoods $\{\score_k\}$}
		\State $\{\score_k,\rcenter_k\}_{k=1}^K=\text{top}(K, \{s^b(\bond_{ij})\}^K,\{s^c_k(\bond_{ij})\}^K,\{s^a(\atom_{i})\}^K)$
		\vspace{0.3em}
		\LineComment{convert a product into $K$ sets of synthons and update their log-likelihoods with Supplementary Algorithm~\ref{alg:encoder}}
		\State $\{\score_k,\graph_{s,k}\}_{k=1}^K = \ours\text{-}\pstrans(\graphp, \{\score_k,\rcenter_k\}_{k=1}^{K}, \hidden_p)$
	    \vspace{0.3em}
		\State \Return  $\{\score_k,\graph_{s,k}\}_{k=1}^K$
	\end{algorithmic}
\end{algorithm}

\begin{algorithm}[!h]
	\caption{\ours-encoder}
	\label{alg:encoder}
	\begin{algorithmic}[1]
		\Require \graph, \tree
		\vspace{0.3em}
		\LineComment{calculate atom embeddings}
		\State $\{\atomEmb_i\}=\GMPN(\graph)$
		\vspace{0.3em}
		\LineComment{calculate the graph embedding (Equation~\ref{eqn:pool})}
		\State $\hidden=\sum_{\scriptsize{\atom_i\in\graph}}\atomEmb_i$
		\vspace{0.5em}
		\If{use \brics}
		\State $\{\nodeEmb_u\} = \BMPN(\tree, \{\atomEmb_i\})$
		\ShortLineComment{update the embedding of each atom with its \brics fragment embedding (Equation~\ref{eqn:enrich})}
		\State $\{\atomEmb_i\} = \{V(\atomEmb_i\oplus\nodeEmb_u)\}$
		\EndIf
		\vspace{0.3em}
		\LineComment{calculate bond embeddings (Equation~\ref{eqn:bondEmb})}
		\State $\{\bondEmb_{ij}\}=\text{bondEmb}(\{\atomEmb_i\})$
		\vspace{0.3em}
		\State \Return  $\{\atomEmb_i\}$, $\{\bondEmb_{ij}\}$, \hidden
	\end{algorithmic}
\end{algorithm}

\begin{algorithm}[!h]
	\caption{\ours-\pstrans for transformation from product to synthons}
	\label{alg:p2st}
	\begin{algorithmic}[1]
		\Require $\graphp$, $\{\score_k,\rcenter_k\}_{k=1}^K$, $\hidden_p$
		\vspace{0.3em}
		\For{each $\score_k$, $\rcenter_k$}
		\vspace{0.3em}
		\If{$\rcenter_k$ is \bondfmC}
		\ShortShortLineComment{predict bonds with induced type changes and calculate the log-likelihood $\score_{\mathsf{BF}}$ for the predictions of $\neighCSet(\rcenter_k)$}
		\State $\neighCSet^\prime(\rcenter_k), \score_{\mathsf{BF}}=\BTCP(\rcenter_k, \neighCSet(\rcenter_k), \hidden_{p})$
		\ShortShortLineComment{add the predicted bonds with type changes into the center}
		\State $\rcenter_k=\rcenter_k \cup \neighCSet^\prime(\rcenter_k)$
		\ShortShortLineComment{update the log-likelihood score}
		\State $\score_k = \score_k + \score_{\mathsf{BF}}$
		\EndIf
		\vspace{0.3em}
		\ShortLineComment{predict atoms with charge changes for all the atoms within the center $\neighACSet(\rcenter_k)$, and calculate the log-likelihood score $\score_{\mathsf{A}}$ for the predictions of $\neighACSet(\rcenter_k)$}
		\State $\neighACSet^\prime(\rcenter_k), \score_{\mathsf{A}} = \ACP(\rcenter_k, \neighACSet(\rcenter_k), \hidden_{p})$
		\ShortLineComment{update the log-likelihood score}
		\State $\score_k = \score_k + \score_{\mathsf{A}}$
		\vspace{0.3em}
		\ShortLineComment{transform the product graph into the synthon graph with reaction center $\rcenter_k$ and atom charge change $\neighACSet^\prime(\rcenter_k)$}
		\State $\graph_{s,k} = \text{transform}(\graphp, \rcenter_k, \neighACSet^\prime(\rcenter_k))$
		\EndFor
		\vspace{0.3em}
		\State \Return  $\{\score_k, \graph_{s,k}\}_{k=1}^K$
	\end{algorithmic}
\end{algorithm}

\begin{algorithm}[!h]
	\caption{\ours-\syncomp for synthon completion}
	\label{alg:syncomp}
	\begin{algorithmic}[1]
		\Require $\graphp$, $\{\score_k,\graph_{s,k}\}_{k=1}^K$, $N$, maxSteps
		\vspace{0.3em}
		\State $t=0$
		\vspace{0.3em}
		\LineComment{learn molecule representations with Supplementary Algorithm~\ref{alg:encoder}}
		\State -, -, $\hidden_p = \ours\text{-encoder}( \graphp )$
		\State -, -, $\{\hidden_{s,k}\}_{k=1}^K = \ours\text{-encoder}(\{\graph_{s,k}\}_{k=1}^K)$
		\vspace{0.3em}
		\LineComment{initialize a priority queue with synthons $\{\graph_{s,k}\}_{k=1}^K$ as elements and $\{\score_k\}_{k=1}^K$} as their priorities
		\State $Q^{(0)}=\text{priorityQueue}(\{\score_k,\graph_{s,k}\}_{k=1}^K)$
		\vspace{0.3em}
		\LineComment{initialize an empty priority queue to store complete reactants}
		\State $R = \text{priorityQueue}()$
		\vspace{0.3em}
		\While{$!Q^{(t)}.\text{isEmpty()}$ and $t\leq$ maxSteps}
		\vspace{0.3em}
		\ShortLineComment{stop the completion when it is impossible to get reactants better than the top-$N$ reactants in $R$}
		\If{$R.\text{size()} \geq N$ and $R.\text{nthLargestPriority}(N)\geq Q^{(t)}.\text{maxPriority()}$}
		\State \textbf{break}
		\EndIf
		\vspace{0.3em}
		\ShortLineComment{complete synthons through beam search with Supplementary Algorithm~\ref{alg:beamsearch}}
		\State $Q^{(t+1)}, R = \ours\text{-beam-search}(Q^{(t)}, R, \{\hidden_{s,k}\}_{k=1}^K, \hidden_p, N)$
		\State $t = t + 1$
		\EndWhile
		\vspace{0.3em}
		\LineComment{output top-$N$ reactants}
		\State $\{\graph_{r,i}\}^N_{i=1} = R.\text{nLargest}(N)$
		\vspace{0.3em}
		\State \Return  $\{\graph_{r,i}\}^N_{i=1}$
	\end{algorithmic}
\end{algorithm}

\begin{algorithm}[!h]
	\caption{\ours-beam-search}
	\label{alg:beamsearch}
	\begin{algorithmic}[1]
		\Require $Q$, $R$, $\{\hidden_{s,k}\}_{k=1}^K$, $\hidden_p$, $N$
		\vspace{0.3em}
		\State $I = Q.\text{size}()$
		\vspace{0.3em}
		\State $Q^\prime = \text{priorityQueue}()$
		\vspace{0.3em}
		\While{!Q.isEmpty()}
		\vspace{0.3em}
		\State $\score_i, \graph^{*}_i = Q.\text{pop}()$
		\vspace{0.3em}
		\ShortLineComment{get the index of the synthon corresponding to $\graph^{*}_i$}
		\State $k = \graph^{*}_i.\text{getSynthonIdx}()$
		\vspace{0.3em}
		\ShortLineComment{predict the atom attachment for $\graph^{*}_i$ with Supplementary Algorithm~\ref{alg:aap}}
		\State $\{\score_{i,j}, \graph_{i,j}^{\prime}\}_{j=1}^{N+1}=\ours\text{-AAP}(\graph^{*}_i, \score_i, \hidden_{s,k}, \hidden_p, N)$
		\vspace{0.3em}
		\EndWhile
		\vspace{0.3em}
		\LineComment{select top-$N$ intermediate graph candidates}
		\State $\{\score_{i}, \graph_{i}^{\prime}\}^{N}_{i=1} = \text{top}(N, \{\{\score_{i,j}, \graph_{i,j}^{\prime}\}^{N+1}_{j=1}\}^{I}_{i=1})$
		\vspace{0.3em}
		\For{each $\score_{i}$, $\graph_{i}^{\prime}$}
		\vspace{0.3em}
		\If{$\graph_{i}^{\prime}$.isComplete()}
		\State $R$.push($\score_{i}$, $\graph_{i}^{\prime}$)
		\Else
		\State $Q^\prime$.push($\score_{i}$, $\graph_{i}^{\prime}$)
		\EndIf
		\EndFor
		\vspace{0.3em}
		\State \Return  $Q^\prime$, $R$
	\end{algorithmic}
\end{algorithm}

\begin{algorithm}[!h]
	\caption{\ours-\AAP for atom attachment prediction}
	\label{alg:aap}
	\begin{algorithmic}[1]
		\Require $\graph^*$, $\score$, $\hidden_s$, $\hidden_p$, $N$
		\vspace{0.3em}
		\LineComment{get the atom that new substructures will be attached to}
		\State $\atom = \graph^*.\text{nextAttachmentPoint()}$
		\vspace{0.3em}
		\LineComment{predict whether further attachment should be added to \atom (Equation~\ref{eqn:nfscore})}
		\State $\score^o, \score^{\neg o}=\AACP(\atom, \hidden_s, \hidden_p)$
		\vspace{0.3em}
		\LineComment{extend $\graph^{*}$ to the candidate $\graph^\prime_{1}$ that is predicted to stop at \atom}
		\State $\graph^\prime_{1} = \text{stop}(\graph^{*}, \atom)$
		\vspace{0.3em}
		\LineComment{update the log-likelihood value of $\graph^\prime_{1}$}
		\State $\score^\prime_{1} = \score + \score^{\neg o}$
		\vspace{0.3em}
		\LineComment{predict the top-$N$ new substructure attachments (Equation~\ref{eqn:fragpred})}
		\State $\{\frag_i, \score^{\scriptsize{\frag}}_i\}^N_{i=1} = \text{top}(N, \AATP(\atom, \hidden_s, \hidden_p))$
		\vspace{0.3em}
		\LineComment{extend $\graph^{*}$ to the candidates $\{\graph^\prime_{i}\}^{N+1}_{i=2}$ with the top-$N$ substructures}
		\State $\{\graph^\prime_{i}\}^{N+1}_{i=2}=\text{attach}(\graph^{*},\{\frag_i\}^N_{i=1})$
		\vspace{0.3em}
		\LineComment{update the log-likelihood values of $\{\graph^\prime_{i}\}^{N+1}_{i=2}$}
		\State $\{\score^{\prime}_i\}^{N+1}_{i=2}=\{\score + \score^{o} + \score^{\scriptsize{\frag}}_i\}^N_{i=1}$
		\vspace{0.3em}
		\State \Return  $\{\score^{\prime}_i, \graph^\prime_i\}^{N+1}_{i=1}$
	\end{algorithmic}
\end{algorithm}

\FloatBarrier

\section{Notations}
\label{appendix:notation}

\begin{table}[!h]
    \caption{\textbf{Notations}}
  \label{tbl:notation}
  \centering
  \begin{threeparttable}
      \begin{tabular}{
	@{\hspace{2pt}}l@{\hspace{4pt}}
	@{\hspace{4pt}}l@{\hspace{2pt}}          
	}
        \toprule
        Notation & Meaning \\
        \midrule
        \molr/\mols/\molp                                  & reactant/synthon/product molecule \\
        $\graph=(\atoms, \bonds)$                & molecular graph with atoms \atoms and bonds \bonds \\
        \atom                                    & an atom in \graph \\
        $\bond_{ij}$                                    & a bond in \graph connecting $\atom_i$ and $\atom_j$\\
        $\vect{x}$                               & a feature vector for an atom or a bond \\
        \neighCSet                               & a set of bonds neighboring the bond formation center \\
        \neighACSet                              & {a set of atoms within the reaction center} \\
        \frag                                    & a substructure used to complete synthons into reactants \\
        \vocab                                   & a vocabulary with all the substructures in the dataset\\
         \bottomrule
      \end{tabular}
  \end{threeparttable}
  \vspace{-5pt}
\end{table}

\FloatBarrier
\section{Substructures used to complete synthons}
\label{appendix:subst}

\begin{figure*}[!h]
	\centering
	\includegraphics[width=0.9\linewidth]{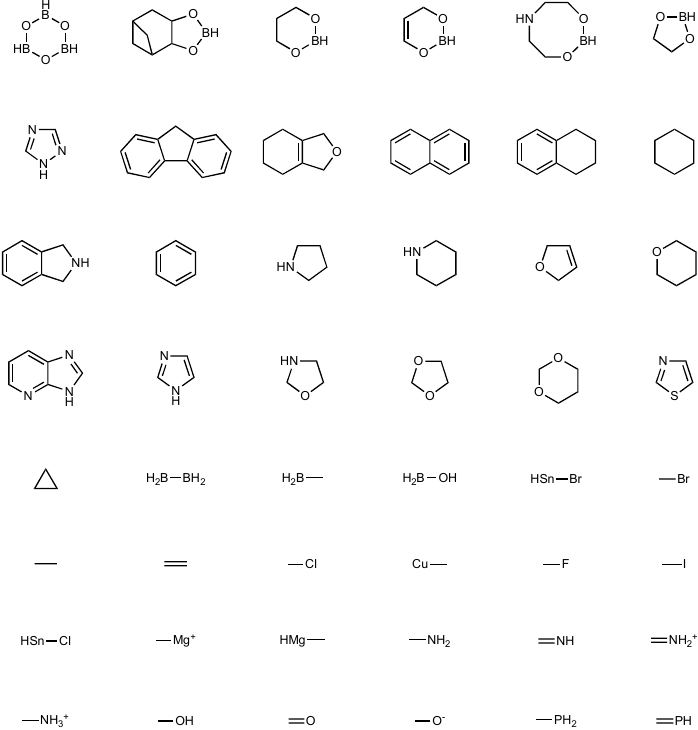}
	\caption{\textbf{83 substructures that \ours uses to complete synthons}}
	\label{fig:str}
\end{figure*}

\begin{figure*}[!h]
	\centering
	\includegraphics[width=0.9\linewidth]{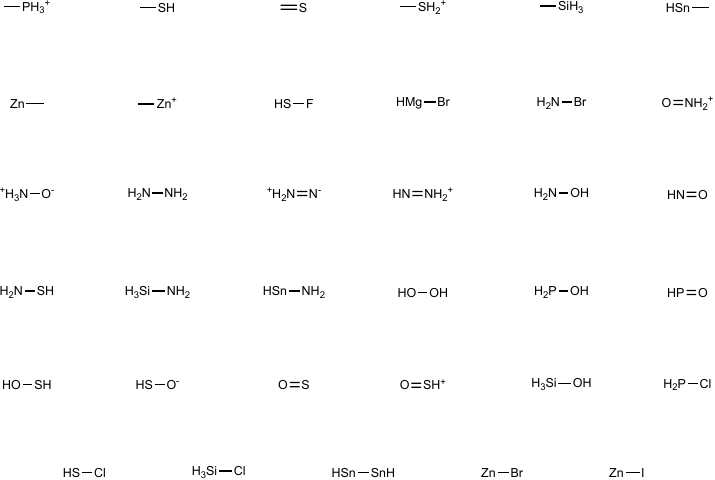}
	\caption{\textbf{83 substructures that \ours uses to complete synthons}}
	\label{fig:str}
\end{figure*}

\section{Parameters for Reproducibility}
\label{appendix:reproducibility}

We tuned the hyper-parameters of the reaction center
identification module and the synthon completion module for \ours and \oursb with the grid-search algorithm.
We presented the parameter space in Supplementary Table~\ref{tbl:parameter}.
We determined the optimal hyper-parameters of the two modules for \ours and \oursb according to the 
corresponding top-1 accuracy over the validation molecules.

\begin{table*}[!h]
    \caption{\textbf{Hyper-parameter space for \ours and \oursb}}
  \label{tbl:parameter}
  \centering
  \begin{small}
  \begin{threeparttable}
      \begin{tabular}{
	@{\hspace{2pt}}l@{\hspace{4pt}}
	@{\hspace{4pt}}l@{\hspace{2pt}}          
	}
        \toprule
        Hyper-parameters & Values \\
        \midrule
        hidden layer dimension & \{128, 256, 512\} \\
        atom embedding dimension & \{32\} \\
        \# iterations of \GMPN  & \{5, 7, 10\} \\
        \# iterations of \BMPN in \oursb & \{3, 5, 7\} \\
         \bottomrule
      \end{tabular}
  \end{threeparttable}
  \end{small}
  \vspace{-5pt}
\end{table*}

In the reaction center identification module, when reaction types are known, the optimal hidden dimension for \ours and \oursb 
is 512; the optimal iterations of \GMPN for \ours and \oursb are 7 and 10, respectively; the optimal iteration of \BMPN for \oursb is 7.
When reaction types are unknown, the optimal hidden dimension for \ours and \oursb is 512 and 256, respectively; the 
optimal iterations of \GMPN for \ours and \oursb are 5 and 10, respectively; the optimal iteration of \BMPN for \oursb is 7.
\ours and \oursb share the same synthon completion model.
In the synthon completion module, when reaction types are known, the optimal hidden dimension is 512; the optimal iteration of \GMPN for \ours is 5. 
When reaction types are unknown, the optimal hidden dimension is 512; the optimal iterations of \GMPN is 7. 

We optimized the models with batch size 256, learning rate 0.001 and learning rate decay 0.9. 
For the reaction center module, we trained the models for 150 epochs and checked the validation accuracy at the end of 
each epoch. We reduced the learning rate by 0.9 if the validation accuracy does not increase by 0.01 for 10 epochs.
We saved the model with the optimal top-3 accuracy on reaction center identification over the validation dataset.
For the synthon completion module, we trained the models for 100 epochs and checked the validation accuracy at the 
end of each epoch over 2,000 reactions that are randomly sampled from the validation set.
We reduced the learning rate by 0.9 if the validation accuracy does not increase by 0.01 for 5 epochs.
We saved the model with the optimal top-1 accuracy on synthon completion over the sampled subset of the validation dataset.

We implemented our models using Python-3.6.9, Pytorch-1.3.1, RDKit-2019.03.4 and NetworkX-2.3. 
We trained our models on a Tesla P100 GPU and a CPU with 32 GB memory on Red Hat Enterprise 7.7.
The training of our reaction center identification model took $16\sim18$ hours, while the training of our synthon
completion model took $32\sim34$ hours.


%

\end{document}